\documentclass[format=acmlarge, review=false, screen=true]{acmart}
\renewcommand\footnotetextcopyrightpermission[1]{} 
\usepackage{xpatch}
\usepackage{color}
\usepackage{booktabs} 
\usepackage{amsmath,amssymb}
\usepackage[ruled]{algorithm2e} 
\usepackage{graphicx}
\SetAlFnt{\small}
\SetAlCapFnt{\small}
\SetAlCapNameFnt{\small}
\SetAlCapHSkip{0pt}
\IncMargin{-\parindent}
\usepackage{mathrsfs}

\newcommand{\amax}[1]{\text{argmax}_{#1}}
\def\E{\mathbb{E}} 
\def\R{\mathbb{R}} 
\def\P{\mathbb{P}} 
\def\bx{{x}} 
\def\ba{{a}} 
\def\tx{\tilde{\bx}}
\def\Rc{\mathscr{R}} 
\def\Rm{\mathscr{R}^m} 
\def\F{\mathscr{F}} 
\def\Fn{\mathscr{F}^n} 
\def\VoI{\text{VoI}} 
\def\Cov{\text{Cov}\,}

\def\f{{objective function }} 
\def\fNospace{{objective function}}
\def\X{{solution space }}
\def\x{{solution }}
\def\xs{{solutions }}
\def\xr{{recommended solution}}
\def\A{{parameter space }}
\def\a{\text{parameter }}
\def\as{{parameters }}
\def\aNospace{\text{parameter}}
\def\asNospace{{parameters}}

\def\B{{B }}

\def\xa{{simulation point }}
\def\xaNospace{{simulation point}}
\def\xas{{simulation points }}
\def\xasNospace{{simulation points}}
\def\y{{performance }}

\def\T{{target function }}

\def\s{{parameter data source }}
\def\sNospace{{parameter data source}}
\def\ss{{parameter data sources }}
\def\ssNospace{{parameter data sources}}
\def\r{{data source sample }}
\def\rs{{data source samples }}
\def\rsNospace{{data source samples}}
\def\Pa{\text{parameter distribution}}
\def\sr{\text{queried data pair}}
\def\srs{\text{queried data pairs}}

\def\Dm{\mathscr{R}^{m}}
\newtheoremstyle{mystyle}
  {}
  {}
  {\itshape}
  {}
  {\bfseries}
  {.}
  { }
  {}

\theoremstyle{mystyle}

\newtheorem{thm}{Theorem}
\newtheorem{ppr}{Proposition}
\newtheorem{remark}{Remark}
\begin{document}
	\title{Bayesian Optimisation vs.\ Input Uncertainty Reduction}
	

	\author{Juan Ungredda}
	\affiliation{%
	  \institution{University of Warwick}
	  \department{Complexity Science Centre}
	  \streetaddress{Gibbet Hill Road}
	  \city{Coventry}
	  \postcode{CV4 7AL}
	  \country{UK}}
	\email{m.a.l.pearce@warwick.ac.uk}
	
	\author{Michael Pearce}
	\affiliation{%
		\institution{University of Warwick}
		\department{Mathematics for Real-World Systems Centre for Doctoral Training}
		\streetaddress{Gibbet Hill Road}
		\city{Coventry}
		\postcode{CV4 7AL}
		\country{UK}}
	\email{J.Ungredda@warwick.ac.uk}
	
	\author{Juergen Branke}
	\affiliation{%
		\institution{University of Warwick}
		\department{Warwick Business School}
		\streetaddress{Gibbet Hill Road}
		\city{Coventry}
		\postcode{CV4 7AL}
		\country{UK}}
	\email{juergen.branke@wbs.ac.uk}
	
	\begin{abstract}
	Simulators often require calibration inputs estimated from real world data
	and the quality of the estimate can significantly affect simulation output.
	Particularly when performing simulation optimisation to find an optimal solution,
	the uncertainty in the inputs significantly affects the quality of the found solution. 
	One remedy is to search for the solution that has the best performance on average
	over the uncertain range of inputs yielding an optimal compromise solution. 
	We consider	the more general setting where a user may choose between either running simulations or instead
	collecting real world data. A user may choose an input and a solution and observe the simulation output, or instead query an external data source improving the input estimate enabling 
	the search for a more focused, less compromised solution. We explicitly examine the trade-off between
	simulation and real data collection in order to find the optimal solution of the simulator 
	with the true inputs.
%
%
	Using a value of information procedure, we propose a novel unified simulation optimisation procedure
	called Bayesian Information Collection and Optimisation (BICO) that, in each iteration, automatically determines which of the two actions (running simulations or
	data collection) is more beneficial. Numerical experiments demonstrate that the proposed algorithm
	is able to automatically determine an appropriate balance between optimisation and data collection.
	\end{abstract}

	\keywords{Input Uncertainty, Simulation Optimisation,
		Gaussian Processes, Bayesian Optimisation}

	\maketitle

\section{Introduction}

Simulators are often used as cheap surrogate models of real world systems, enabling users to prototype and test possible solutions before deploying such a solution in practice. Simulation optimisation is the problem of identifying the best solution, when solution qualities can only be estimated via sampling, i.e., running a computationally expensive simulation and obtaining a stochastic output value. In many cases, the simulation model has additional parameters that need to be set, such as the mean arrival rate of customers or the mean and variance of the demand distribution.

In reality, such input parameters are either chosen by expert opinion or set to values estimated from historical data.
If the chosen values for the input parameters differ significantly from the true parameters, the solution found by optimising the simulation model may be far from optimal in the real world. 
This problem is generally known as simulation optimisation with input uncertainty and has received much attention in recent years (\citeauthor{lam216}, \citeyear{lam216}). Much work has focused on explicitly modeling the uncertainty of the input parameters and seeking 
a robust solution that performs well on average (or worst case) over this distribution. 

In this paper, we extend our previous work on Bayesian optimisation aiming to identify the solution with the best expected performance given the input uncertainty (\citeauthor{IUbrankepearce}, \citeyear{IUbrankepearce}).
In particular, we assume that the user has access to real world data that can help to inform the parameters required by the simulator. Given finite resources to spend on simulation and/or data collection, an algorithm must carefully determine which of the two possible actions to perform.

Devoting too much effort to data collection may not leave sufficient resources for optimisation and an algorithm would return a sub-optimal solution to an accurate simulator. On the other hand, devoting too little effort to data collection may lead to learning a good \emph{compromise} solution that performs well on average across a variety of possible input parameters, but may be sub-optimal under the true input parameters.
In this work, we propose a Bayesian optimisation algorithm that can intelligently trade off simulation and data collection.

This applies to simulation optimisation problems where extra external input data can be collected incrementally 
requiring resources to collect. For example, manually labelling or cleaning data, sales demand may be estimated from physical sales
records that needs to be manually sorted and entered into a database to reduce uncertainty about true demand.
Alternatively, external data may require time consuming physical measurements by real-world observers such as traffic flow or user choices.

We start with an overview of related work in Section~\ref{sec:LITERATURE REVIEW}, followed by a formal definition of the problem in Section\ref{sec:PROBLEMFORMULATION}. Section~\ref{sec:MODELS} explains the statistical models and Section \ref{sec:ACQUISITION} derives the sampling procedures, their theoretical properties and practical computation. We perform numerical experiments in Section \ref{sec:NUMERICAL}. Finally, the paper concludes with a summary and some suggestions for future work in Section \ref{sec:CONCLUSION}.

 	\section{LITERATURE REVIEW}\label{sec:LITERATURE REVIEW} 
Bayesian optimisation (BO) builds a Gaussian process, or Kriging, surrogate model of the simulator response surface based on a few initial samples and then uses an acquisition function, or infill criterion, to sequentially decide where to sample next in order to improve the model and find
better solutions. For a brief introduction refer to \citeauthor{NandoDeFreitas} (\citeyear{NandoDeFreitas}). 

Several BO algorithms have been proposed in the literature. The most popular is the Efficient Global Optimisation (EGO) algorithm of \citeauthor{Jones198} (\citeyear{Jones198}) that combines a Gaussian Process to interpolate an expensive function with an expected improvement criterion for deciding where to sample next. The Knowledge Gradient (KG) policy for Continuous Parameters (\citeauthor{Frazier211} \citeyear{Frazier211}) is another myopic sampling policy that aims to maximise the new predicted optimal performance after one new sample. Different from EGO, KG accounts for covariance when judging the value of a sample and can be directly applied to noisy functions.

Conventional optimisation approaches, including BO, assume that the auxiliary input parameters are known when often this is not the case. Therefore, investigating the effect of input uncertainty has recently gained significant interest in the simulation community, for a general introduction see, e.g., \citeauthor{lam216} (\citeyear{lam216}). Currently, there are several proposed methods to assess the input uncertainty and its impact on the mean value of the simulation output. \citeauthor{BartonS01} (\citeyear{BartonS01}) built an empirical distribution given historical data and sample from it using direct and bootstrap techniques to assess the impact of input uncertainty. \citeauthor{Chick&Steph} (\citeyear{Chick&Steph}) uses a  Bayesian posterior distribution to estimate the input distributions for the same purpose. \citeauthor{ChengHolland} (\citeyear{ChengHolland}) estimate the simulation uncertainty through its decomposition into random variations of the simulation model (simulation uncertainty) and the input parameter uncertainty. \citeauthor{BartonRusselNelsonBarryXie} (\citeyear{BartonRusselNelsonBarryXie}) replace the expensive simulation by metamodel-assisted
bootstrapping using a stochastic Kriging response surface  to estimate the impact of input uncertainty on the simulation output. 

The aforementioned methods assume given data to assess the uncertainty. In the case when additional input
data can be collected, \citeauthor{QuicklyAsses} (\citeyear{QuicklyAsses}) propose to consider the relative contributions and
sensitivities to the overall effect of input uncertainty to give guidance about the best inputs to update. \citeauthor{FreiSchr02}(\citeyear{FreiSchr02}) examine the question how much data to collect, and for what parameters. They suggest to run an initial experimental design with the endpoints of the confidence interval of the input uncertainty. Then they can use ANOVA to see whether the parameter effects are significant. If they are, then more information should be collected to reduce the uncertainty of the parameter.  For a simplified setting only considering main effects, \citeauthor{SongNelson15} \citeyear{SongNelson15} propose a more efficient method that 
approximates the impact of input uncertainty on the overall variance in the simulation output with the help of a mean-variance metamodel depending on the means and variances of the input distributions. They suggest using the resulting sensitivities for deciding which additional data to collect.

When input uncertainty estimation is considered in the optimisation process, \citeauthor{E.Song215} (\citeyear{E.Song215}) explore the impact of model risk due to input uncertainty on indifference zone (IZ) ranking \& selection. \citeauthor{WuZhou} (\citeyear{WuZhou}) use ranking and selection in a two-stage allocation of finite budget, where the first stage consists in estimating the input parameters, followed by the budget allocation scheme to perform simulation runs in the second stage. \citeauthor{XiaoGao} (\citeyear{XiaoGao}) consider taking the input uncertainty into account, but the optimisation is focused on the worst-case performance given a fixed finite number of input models. \citeauthor{ZhouXie} (\citeyear{ZhouXie}) propose a formulation that allows to adapt to one's risk preference for the optimisation.

Only very few papers consider the case where additional information can be gathered during the optimisation process.
\citeauthor{SongShanbhag2019} (\citeyear{SongShanbhag2019})  consider the case of optimisation under input uncertainty when additional data is received from an uncontrolled streaming data process during optimisation. They propose a stochastic approximation framework that prescribes the number of gradient descent steps to be conducted in every time step.
For the discrete ranking and selection problem, \citeauthor{WuZhou2019} (\citeyear{WuZhou2019}) study the impact of input uncertainty assuming new data becomes available in each iteration. They propose a technique that discards the oldest simulation outputs in the estimation of the means and an elimination of designs according to its confidence bounds. They
propose a stopping criterion that has a guaranteed probability of correct selection.

In this work, we explicitly look at the trade off between running more simulations or input data collection with the aim of finding the optimal solution to a simulator with accurate input parameters.
Our methodology builds on previous work by \citeauthor{IUbrankepearce} (\citeyear{IUbrankepearce}) who extended Efficient Global Optimization
(EGO) and Knowledge Gradient (KG) with Continuous Parameters so that they work efficiently under input
uncertainty. In broader terms, this problem can be described as optimising an integrated expensive-to-evaluate
function (\citeauthor{toscanopalmerin2018bayesian} \citeyear{toscanopalmerin2018bayesian}). A similar extension
has been proposed for the Informational Approach to Global Optimization (IAGO) algorithm by \citeauthor{WangYuanNg} (\citeyear{WangYuanNg}).


\section{PROBLEM FORMULATION}\label{sec:PROBLEMFORMULATION}

For simulation data, we assume \textit{\xs} are given by vectors in a \textit{\X}, $\bx \in X \subset \R^D$. The simulator may have multiple inputs for different purposes and we refer to the concatenated vector as \textit{\as} in \textit{\A}, $\ba \in A \subset \R^J$. The simulator is an arbitrary black box we refer to as the \textit{\f}
$$
f: X \times A \to \R
$$
which takes as arguments a solution and parameters and returns a noisy scalar valued \textit{\y} $y$. Finally, the expectation of noisy performance is referred to as the \textit{\T} denoted
$\theta(\bx,\ba) = \E[f(\bx,\ba)]$.

For \a data collection, we let $N_s$ be the number of \textit{\ss} indexed by $s\in S = \{1,...,N_s\}$ (where $N_s$ may or may not equal parameter dimension $J$). Querying a data source $s$ returns a \textit{\a data point} $r\sim \P[r|a^*,s]$ where $\ba^*$ is the true \a vector. $\ba^*$ may be inferred using the likelihood of the data 
\begin{equation} \label{eq:astarlikelihood}
\prod_{i=1}^m\P[r^i|s^i,a]    
\end{equation}
where $m$ is the number of data samples collected so far, and $r^i$ denotes the value observed from data source $s^i$ from which the $i^{th}$ data sample was collected. The likelihood is defined by the application at hand and therefore we
assume it is given and may be used by any algorithm.

\begin{figure}[H]
	\centering
	
	\includegraphics[height=5.5cm, trim={10 0 450 0},clip]{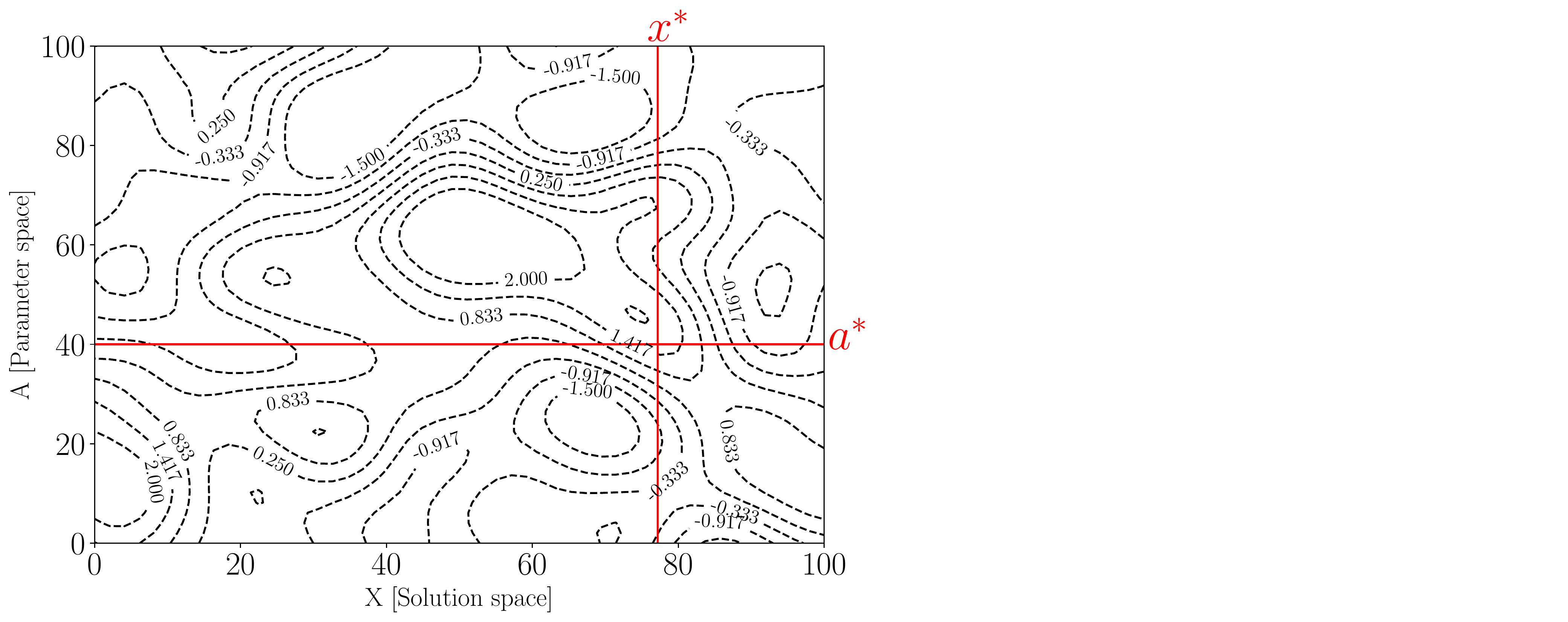}\\

	\captionof{figure}{Surface $\theta (\bx,\ba)$ with true \a $\ba^{*}$ and \x $\bx^{*}$. The goal of an optimisation algorithm is to learn $x^* = \amax \theta(x, a^*)$ which requires learning both the true input parameters $a^*$ as well as the true response surface $\theta(x,a)$ particularly $\theta(x,a^*)$.}
	\label{fig:True_perfvsExpected1}
	
\end{figure}

For the goal of optimisation, both simulation triplets $(\bx,\ba,y)$ and \a data pairs $(s, r)$ must be collected to infer both $\theta(x,a)$ and $a^*$ respectively. The aim is to learn the true best solution 
$$\bx^* = \amax{\bx}\theta(\bx, \ba^*).$$ Figure~\ref{fig:True_perfvsExpected1} illustrates an example.

There is a budget of \B units that can be spent either by choosing $(\bx, \ba)$ and calling $f(\bx,\ba)$ costing $c_f$, or by choosing $s\in\{1,...,N_s\}$ and querying $\P[r|s, a^*]$ costing $c_{s}\in\{c_1$,...,$c_{N_s}\}$. After consuming the budget, a solution $\bx_{r}$ is returned to the user and its quality is determined by the difference in true performance between $\bx_{r}$ and the best solution $\bx^{*}$, or Opportunity Cost (OC),

 \begin{align}\label{eq:OC}
 OC(\bx_{r}) = \theta(\bx^{*}, a^{*})-\theta(\bx_{r},a^{*})
\end{align}

As example, in Section \ref{sec:NUMERICAL} we consider the newspaper vendor problem. A news vendor aims to maximise profit by choosing the optimal number of newspapers to stock. However, the demand for newspapers is uncertain and significantly affects the optimal number of newspapers to stock. We have a newsvendor simulator to evaluate any chosen stock level with any set demand, and we also have access to a supply of past sales. We can collect either more simulator data or more past sales data in order to find to true optimal stock level for the true demand level.

  	\section{The BICO Algorithm}\label{sec:MODELS}
We propose the Bayesian Information Collection and Optimisation algorithm (BICO) that automatically decides whether to conduct additional simulation experiments to find better solutions or to collect additional parameter data to reduce \a uncertainty. In Sections \ref{sec:StatisticalModelSim} and \ref{sec:StatisticalModelInfo} we describe the statistical models for inferring the target function $\theta(x,a)$ and true \as $a^*$, respectively. Section \ref{sec:ACQUISITION} derives the general Value of Information Procedure and Sections \ref{sec:VoISim} and \ref{sec:VoIInfo} apply this to value collecting simulation and collecting parameter data. At each iteration, the action is simply determined by what has the highest value. Together the modelling and automated value based data collection form the BICO algorithm summarised in Algorithm \ref{alg:Bico}. We then prove properties about BICO behaviour in Section \ref{sec:BICOproofs}.

\subsection{Statistical Model for the Target Function}\label{sec:StatisticalModelSim}
Let us denote the $n$-th simulation point by $(\bx,\ba)^{n}$ and \y by $\mathbf{y}^{n} = f(\bx^n, \ba^n)$ and the set of points up to $n$ as $\Fn = \{(x,a,y)^1,\dots,(x,a,y)^n \}$. For convenience, we define the concatenated objective arguments $\tilde{X}^n = \{(x,a)^{1},\dots,(x,a)^{n}\}$ and $\tx = (x,a)$ and vector of outputs $Y^{n} = (y^1,\dots, y^n)$. We propose to use a Gaussian process (GP) to model $\theta(\bx,\ba)$. A Gaussian process is defined by a mean function $\mu^0(\tx):X\times A \to \R$ and a covariance function $k^0(\tx, \tx'):(X\times A) \times (X\times A) \to \R$. Given the objective function dataset $\Fn$, predictions at new locations $(\bx, \ba)$ are given by 

\begin{align}
\begin{split}\label{eq:eq2}
\mathbb{E}[\theta(\bx,\ba)|\Fn] &= \mu^n(\bx,\ba) \\
&=\mu^0(\bx,\ba)-k^0((\bx,\ba),\tilde{X}^n)(k^0(\tilde{X}^n,\tilde{X}^n)+I\sigma)^{-1}(Y^n-\mu^0(\tilde{X}^n))
\end{split}\\
\begin{split}
\Cov[\theta(\bx,\ba),\theta(\bx',\ba')|\Fn] &= k^n((\bx,\ba);(\bx',\ba'))\\
&= k^0((\bx,\ba);(\bx',\ba'))-k^0((\bx,\ba);\tilde{X}^n)(k^0(\tilde{X}^n,\tilde{X}^n)+I\sigma)^{-1}k^0(\tilde{X}^n;(\bx',\ba'))
\end{split}
\end{align}

The prior mean $\mu^0(\bx,\ba)$ is typically set to $\mu^0(x,a)=0$ and the $k^0(\tx, \tx')$
allows the user to encode known properties of the \T $\theta(\bx,\ba)$ such as smoothness
and periodicity. In Section 5, we use the popular squared exponential kernel that assumes
$\theta(x,a)$ is a smooth function such that nearby $(x,a)$ points have similar outputs while widely separated points have unrelated outputs,
\begin{align}\label{kernel}
k^{0}((\bx,\ba);(\bx',\ba')) = \sigma_{0}^{2}\exp\left( \frac{||(\bx,\ba)-(\bx',\ba')||^{2}}{2l_{XA}^{2}}\right) 
\end{align}
where $\sigma_{0} \geq 0$ and $l_{XA} > 0$ are hyper-parameters estimated from the data $\Fn$ by maximum marginal likelihood described in the Appendix. Further details can be found in \citeauthor{Rasmussen06} \citeyear{Rasmussen06}.

\subsection{Statistical Model for the True Parameters}\label{sec:StatisticalModelInfo}
\begin{figure}[H]
	
	\centering
	\begin{tabular}{cc}
		\includegraphics[height=5.0cm, trim={10 0 450 0},clip]{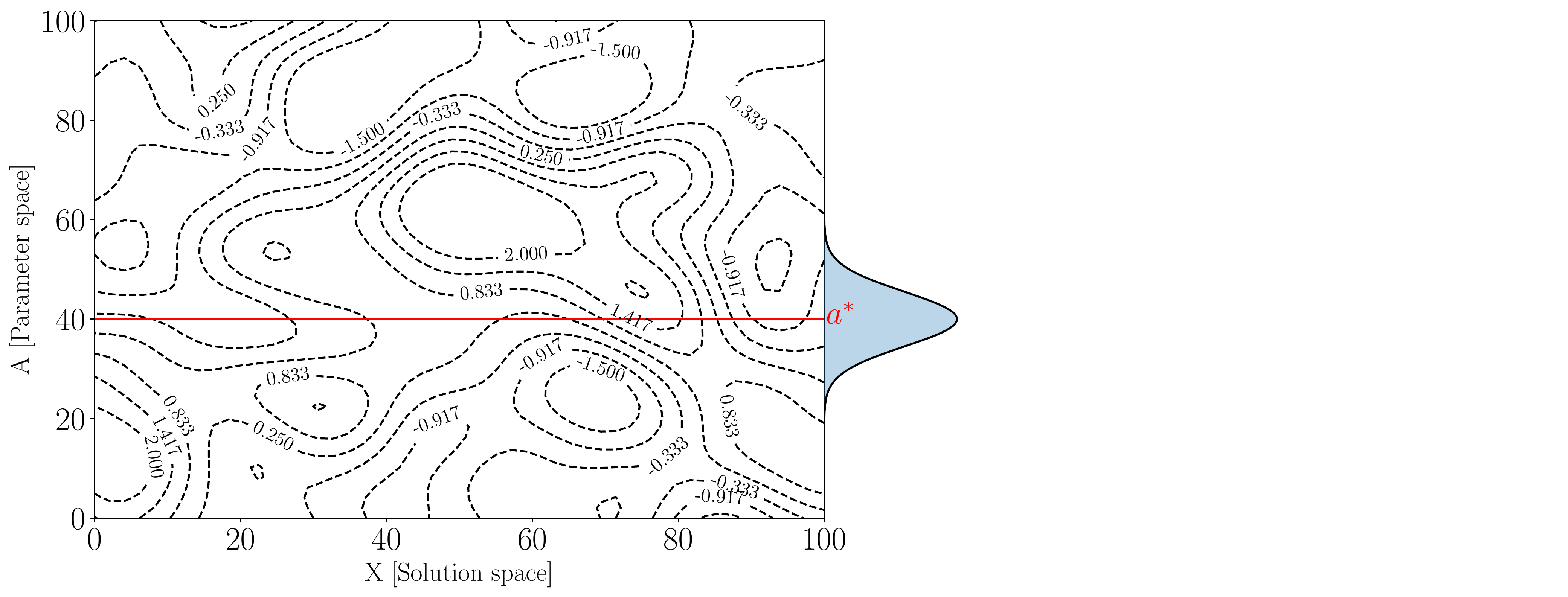}
	\end{tabular}
	\captionof{figure}{Surface $\theta (\bx,\ba)$ sliced by the unknown true parameter $\ba^{*}$ (red), and uncertainty distribution $\P[\ba|\Rm]$ (blue) over the possible input values given collected data.}
	\label{fig:models_acquiring_data}
\end{figure}

We further use a Bayesian approach to estimate the true \a $\ba^{*}$. We denote the set of $m$ \srs 
\\
$\Dm = \{(s, r)^1,....,(s, r)^m\}$. The sources $s^1,..,s^m$ are deterministically chosen by the algorithm and the observed $r^1,...,r^m$ are each independently generated from each corresponding source and have a likelihood given by Equation \ref{eq:astarlikelihood}. In order to supplement data with expert knowledge, we combine this with a prior distribution $\P[a^*]$ resulting in a posterior distribution
$$
\P[a^*| \Dm] \propto \P[a^*]\prod_{i=1}^{m}\P[r^i| a^*, s^i]
$$
By assuming a convenient and intuitive prior distribution, the posterior distribution $\P[\ba|\Rm]$ can be computed analytically and updated as new sources are queried. In this work, we assume a uniform prior $\P[a^*]$ over the box-constrained space $A$ thereby restricting $a^*$ to realistic values. In our experiments in Section \ref{sec:NUMERICAL}, we work with Gaussian 
distributed data $\P[r|s, a^*]$ therefore the posterior $\P[a^*| \Dm]$ is a truncated Gaussian which is analytically tractable. Figure \ref{fig:models_acquiring_data} shows we can evaluate the true \T by taking a slice through the surface $\theta(\bx,\ba=\ba^{*})$. However, we can only estimate a distribution $\P[\ba|\Rm]$ through collected data.

\subsection{Action Space}\label{sec:ACQUISITION}

 At any iteration $t=m+n$, the algorithm can choose a \xa $(\bx,\ba)\in X\times A$ and observe $y=f(\bx,\ba)$, or it may choose a \s $s \in S = \{1,..,N_s\}$ and observe $r \sim \P[r|s, a^*]$. Therefore the set of actions available to the algorithm is $\{X\times A, S\}$. Below we follow the value of information procedure to derive the expected realised benefit of performing a given action, i.e., an acquisition function over the action set. The algorithm, in each iteration, then selects the action with the largest value. 
 
\subsection{Predicted Performance}
First, we consider the output at the end of executing the algorithm. After exhausting the budget $\B$, the algorithm must return a recommended solution $\bx_r$ to the user. 
The \emph{true} value of any given solution $\bx$ is the expected output of the perfect simulator $\theta(\bx, \ba^*)$. However, both $\theta(\bx,\ba)$ and $\ba^*$ are unknown, hence we can make two approximations. Firstly, approximate $\theta(x,a)$ with the GP prediction $\mu^n(x,a)$. Secondly, replace the fixed point $a^*$ with the expectation
over the posterior $\P[a^*|\Dm]$. Thus, the best estimate of true solution $x$ quality, $\theta(x, a^*)$, given the data so far $\Fn, \Dm$ is denoted as $G(\bx; \Rm, \Fn)$ and given by
\begin{equation}\label{eq:eq1}
G(\bx; \Rm, \Fn)= \E_{a}\left[\E[\theta(\bx,\ba)|\Fn]\big|\Dm\right] =\int_{A}\mu^n(\bx,\ba)\P[\ba|\Dm]d\ba.
\end{equation}
Then, the best \x to recommend, $\bx_{r}$, is the solution that maximises the model's current prediction of true output
\begin{align}
    \begin{split}
        x_r(\Rm,\Fn) &= \amax{x} G(\bx; \Rm, \Fn).
    \end{split}\label{x_r}
\end{align}
By using the above $\bx_{r}$, the corresponding predicted true output is the maximum of $G(\cdot)$ which we denote as
\begin{equation}\label{eq:maxG}
    G^*(\Rm, \Fn) = \max_\bx G(\bx; \Rm, \Fn)
\end{equation}
We use $G^*(\Rm, \Fn)$ as the measure of value or quality of the data we currently have. A value of information procedure quantifies the value of an action by computing the one-step look ahead future expectation of this value
and performing the action with maximum future value.

\begin{figure}[H]
	\centering
	\begin{tabular}{c}
		\includegraphics[height=5.0cm, trim={20 0 0 0},clip]{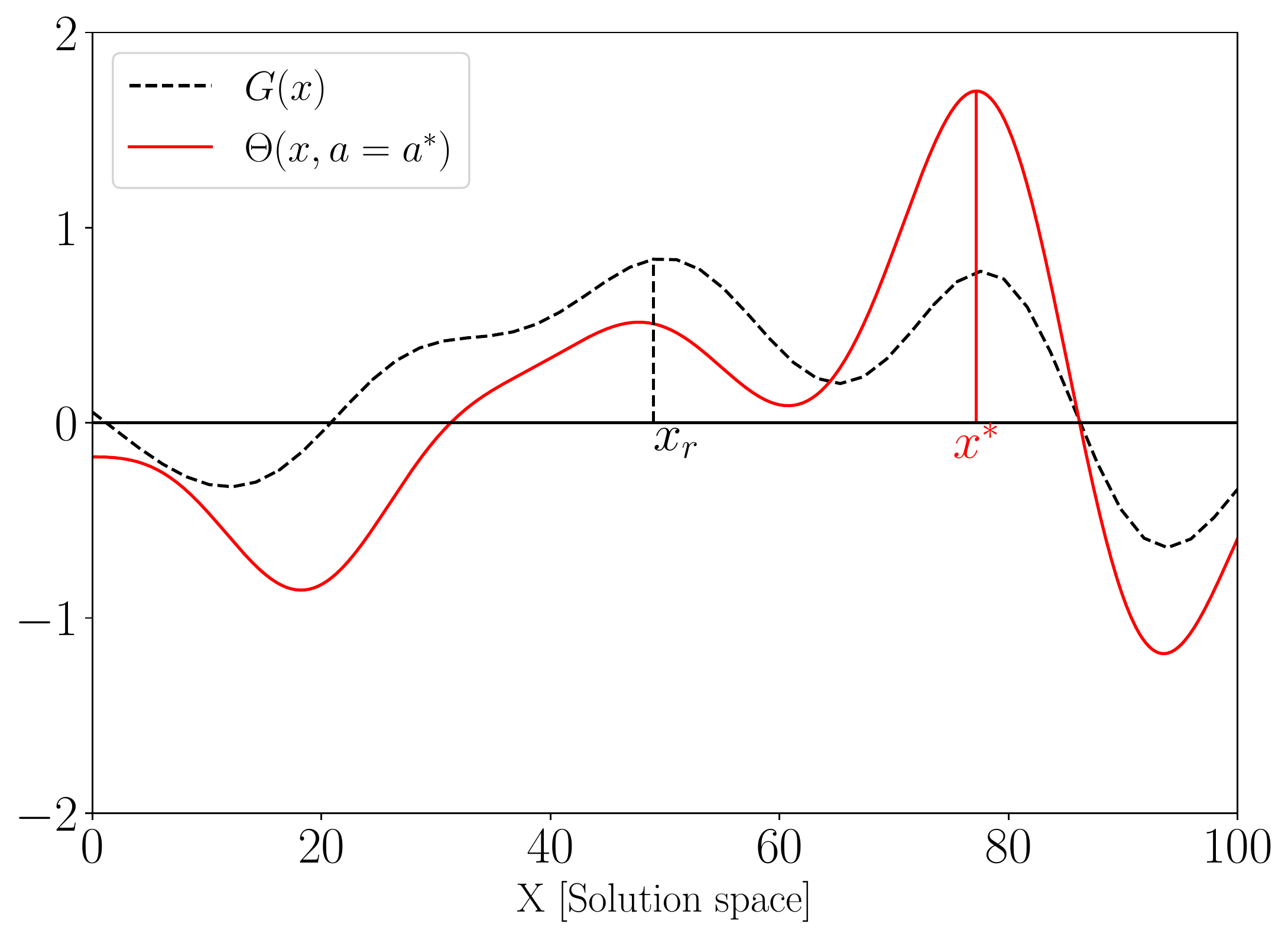}\\
	\end{tabular}
	\captionof{figure}{True \T defined using true \a $\ba^{*}$ (red), and estimated performance using the \Pa ~$\P[\ba|\Rm]$.}
	\label{fig:True_perfvsExpected}
\end{figure}

The difference between using the true \a $\ba^{*}$ and the \Pa ~can be seen in Figure~\ref{fig:True_perfvsExpected}. The predicted solution quality $G(x)$ with the recommended solution $\bx_{r}$ and true quality $\theta(x,a^*)$ with true best \x $\bx^{*}$ may differ substantially. Simulation data helps to improve $\mu^n(x,a)$ to converge towards $\theta(x,a)$. However, even with full simulator information, $\mu^n(x,a)=\theta(x,a)$, the predicted output $G(x)$ must marginalise over $a$ by Equation~\ref{eq:eq1} which is still imperfect and $x_r\neq x^*$.

We next derive the Value of Information (VoI) of performing any action, this is computed by assuming an action is taken and considering the hypothetical predicted performance at the next time step, either $G^*(\Rc^{m+1}, \Fn)$ or $G^*(\Rm, \F^{n+1})$.


\subsection{Value of Information for Simulation Data}\label{sec:VoISim}

If a \xa $(\bx,\ba,y)^{n+1}$  were to be collected thereby augmenting $\F^{n+1} = \Fn \cup \{(\bx,\ba,y)^{n+1}\}$, then the updated predicted performance would be $G( \Rm, \F^{n+1})$. At time $t=m+n$, given the next \xa  $(\bx,\ba)^{n+1}$ and before collecting the new \y $y^{n+1}$, we may compute the one-step look-ahead incremental increase in predicted performance which is the Value of Information (VoI) of taking the action $(\bx,\ba)^{n+1}$,
\begin{equation}
\VoI((\bx,\ba)^{n+1}; \Rm, \Fn) = \E_{y^{n+1}}\left[\frac{ G^*( \Rm, \F^{n+1}) - G^*(\Rm, \Fn)}{c_{f}}\Big| (\bx,\ba)^{n+1} \right] \label{VoI}
\end{equation}
where $c_f$ is the cost of running a simulation. Assuming the datasets are given, $\VoI((\bx,\ba)^{n+1}; \cdot):X\times A\to \R$ is a scalar valued function over the domain of the simulator. It returns the expected increase in simulator output per unit cost of running the simulator.

To evaluate $\VoI((\bx,\ba)^{n+1}; \cdot)$, we next derive the predictive distribution of 
$G(\bx;\Rm, \F^{n+1})$ given data at time $t=n+m$. This requires an updating formula for the posterior mean $\mu^{n+1}(x,a)$. By
setting the posterior mean and covariance after $n$ samples, $\mu^{n}(\bx,\ba)$, $k^{n}((\bx,\ba);(\bx',\ba'))$, as the prior mean and covariance in Eq.~\ref{eq:eq2}, we can write the formula for the mean for the $(n+1)^{th}$ sample as 
\begin{align}
\begin{split}
\mu^{n+1}(\bx,\ba) = \mu^{n}(\bx,\ba) + \frac{k^{n}((\bx,\ba);(\bx,\ba)^{n+1})}{k^{n}((\bx,\ba)^{n+1};(\bx,\ba)^{n+1}) + \sigma^{2}_{\epsilon}}(y^{n+1} - \mu^{n}(\bx,\ba))
\end{split}\label{recurssion8}
\end{align}

where $(\bx,\ba)^{n+1}$ is a given argument to $\VoI(\cdot)$ and $y^{n+1}$ is unknown. The Gaussian Process model provides a predictive distribution for the new function value 

\begin{align}
\begin{split}
y^{n+1} \sim N(\mu^{n}(\bx,\ba)^{n+1},k^{n}((\bx,\ba)^{n+1};(\bx,\ba)^{n+1}) + \sigma^{2}_{\epsilon}).
\end{split}
\end{align}

By writing $y^{n+1} = \mu^n(x,a) + \sqrt{k^{n}((\bx,\ba)^{n+1};(\bx,\ba)^{n+1}) + \sigma^{2}_{\epsilon}}Z$ with $Z\sim N(0,1)$, substituting into Equation \ref{recurssion8} and simplifying leads to the following parametrisation of $\mu^{n+1}(\bx,\ba)$,

\begin{align}
\begin{split}
\mu^{n+1}(\bx,\ba) = \mu^{n}(\bx,\ba) + \tilde{\sigma}^{n}((\bx,\ba);(\bx,\ba)^{n+1})Z 
\end{split}
\end{align}

where $\tilde{\sigma}^{n}((\mathbf{x,a});(\mathbf{x,a})^{n+1})$ is a
deterministic function parametrised by $(\mathbf{x,a})^{n+1}$ that is the additive update to the posterior mean
scaled by $Z$

\begin{align}
\begin{split}
\tilde{\sigma}^{n}((\mathbf{x,a});(\mathbf{x,a})^{n+1}) = \frac{k^{n}((\bx,\ba);(\bx,\ba)^{n+1})}{\sqrt{k^{n}((\bx,\ba)^{n+1};(\bx,\ba)^{n+1}) + \sigma^{2}_{\epsilon}}}
\end{split}
\end{align}

Therefore the predictive distribution of the new posterior mean is given by

\begin{align}
\begin{split}
\mu^{n+1}(\bx,\ba) \sim N(\mu^{n}(\bx,\ba),\tilde{\sigma}^{n}((\mathbf{x,a});(\mathbf{x,a})^{n+1})^{2})
\end{split}
\end{align}

and the predicted performance after a new sample $(\bx,\ba)^{n+1}$ can then be written as

\begin{align}
\begin{split}
G(\bx; \Rm, \F^{n+1}) &= \int_{A}\mu^{n+1}(\bx,\ba)\P[\ba|\Rm]d\ba
\end{split}\\
\begin{split}
&= \int_{A}\mu^{n}(\bx,\ba)\P[\ba|\Rm]d\ba +Z\int_{A}\tilde{\sigma}^{n}((\mathbf{x,a});(\mathbf{x,a})^{n+1})\P[\ba|\Rm]d\ba \label{bigsig}
\end{split}\\
\begin{split}
&= G(\bx; \Rm, \Fn) + Z\tilde{\Sigma}^{n}(\bx;(\bx,\ba)^{n+1})
\end{split}
\end{align}

where $\tilde{\Sigma}^{n}(\bx;(\bx,\ba)^{n+1})$ is the final term in Eq.~\ref{bigsig}. The predictive distribution of a new
observation after evaluating $(\mathbf{x,a})^{n+1}$ is then given by

\begin{align}
\begin{split}
G(\bx; \Rm, \F^{n+1}) \sim N(G(\bx; \Dm, \F^{n}),\tilde{\Sigma}^{n}(\bx;(\bx,\ba)^{n+1})^{2})
\end{split}\label{Gm+1}
\end{align}

The new sample at $\mathbf{(x,a)}^{n+1}$ causes the posterior mean to change at other solutions and inputs according
to the additive update $Z\tilde{\Sigma}^{n}(\mathbf{x};(\mathbf{x,a})^{n+1})$. So, replacing the derived $G(\bx; \Rm, \F^{n+1})$ (Eq. \ref{Gm+1}) in the VoI of acquiring a new \xa $(\bx,\ba)$ (Eq. \ref{VoI}), results in

\begin{align}
\begin{split}
\VoI((x,a)^{n+1}; \Rm, \Fn)&=\frac{1}{c_f}\E_{y^{n+1}}\left[ G^*( \Rm, \F^{n+1})-  G^*( \Rm, \F^{n}) \Big| (\bx,\ba)^{n+1}\right] 
\end{split}\label{KG}\\
\begin{split}
&= \frac{1}{c_f}\mathbb{E}_{z} \Big[\max_{x}\big\{G(\bx; \Rm, \Fn) + Z\tilde{\Sigma}^{n}(\bx;(\bx,\ba)^{n+1})\big\}-G^*(\Rm, \F^{n})\Big|\mathbf{(\bx,\ba)}^{n+1}\Big]
\end{split}\label{VoI_KG}
\end{align}

 The final expectation is identical to the Knowledge Gradient (KG) under input uncertainty with Continuous Parameters (\citeauthor{IUbrankepearce}, \citeyear{IUbrankepearce},  \citeauthor{toscanopalmerin2018bayesian}, \citeyear{toscanopalmerin2018bayesian}). Following these works, the expectation can be evaluated by traditional Knowledge Gradient for Continuous Parameters using Gaussian Processes (\citeauthor{Frazier209}, \citeyear{Frazier209}) where the maximisation over $x\in X$ embedded within the expectation and within $G^*(\cdot)$ are replaced with a maximisation over a disretized set $x\in X_D\subset X$. With this replacement, the expectation over $Z$ can be evaluated analytically. The $\VoI((\bx,\ba); \Rm,\Fn)$ acquisition function may be optimised over the joint solution-input space to find the most beneficial $(\bx,\ba)^{n+1}$ and corresponding $\max\VoI(\cdot)$.

\begin{figure}[htb]
	\centering
	\begin{tabular}{ccc}
		\includegraphics[width=0.31\linewidth]{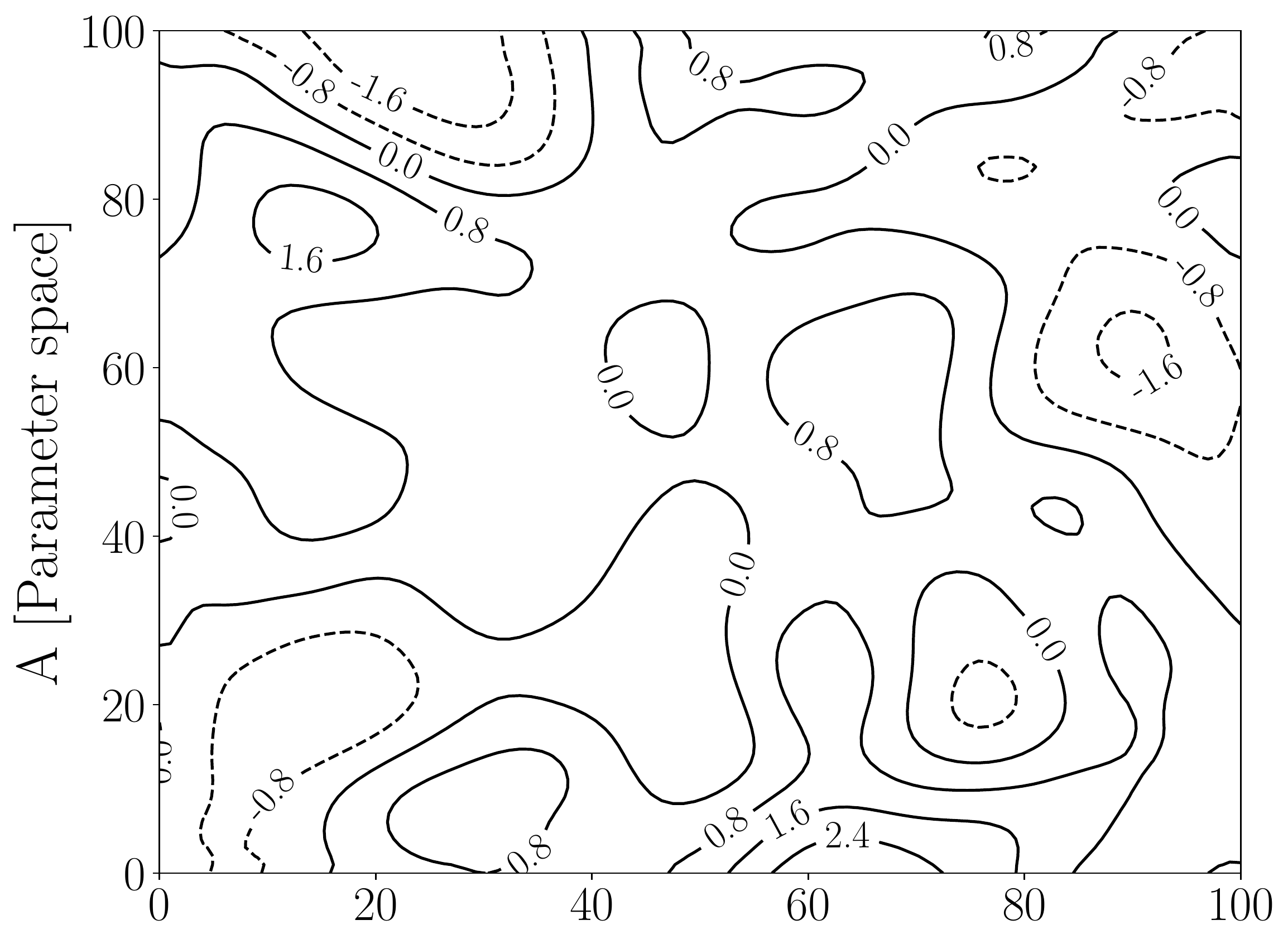}&
		\includegraphics[width=0.31\linewidth]{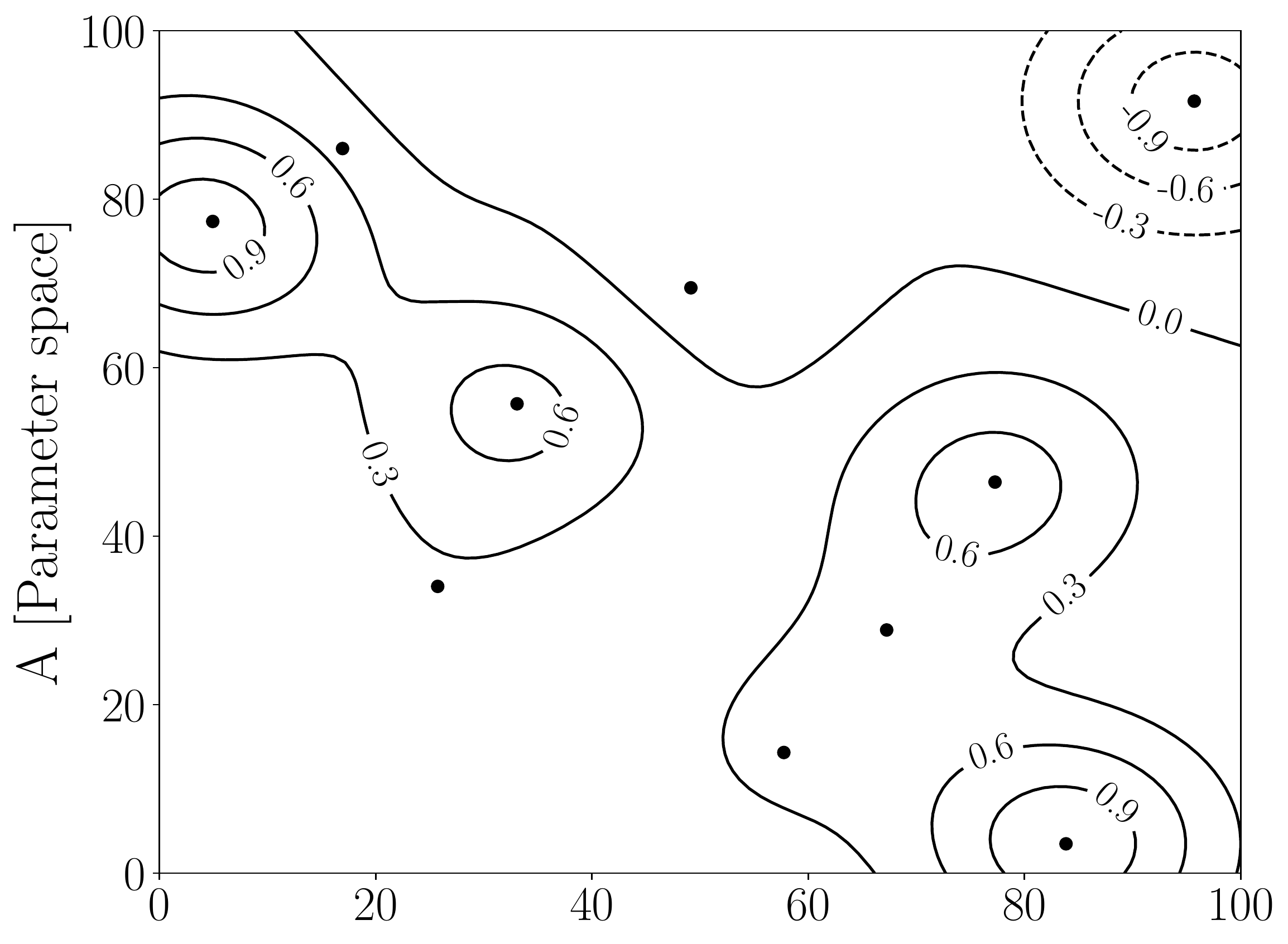}&
		\includegraphics[width=0.31\linewidth]{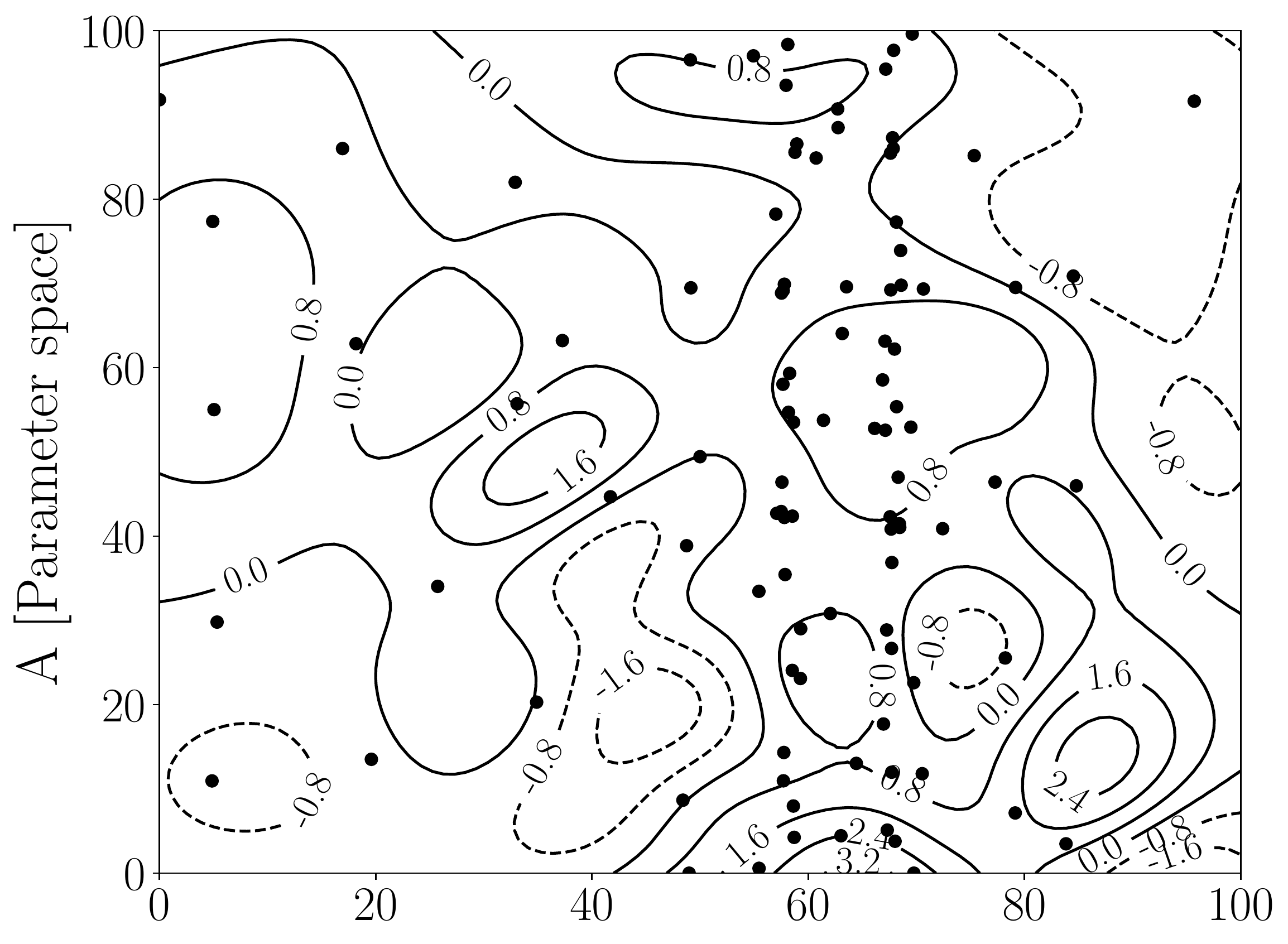}\\
		(a)&(b)&(c)\\
		\includegraphics[width=0.31\linewidth]{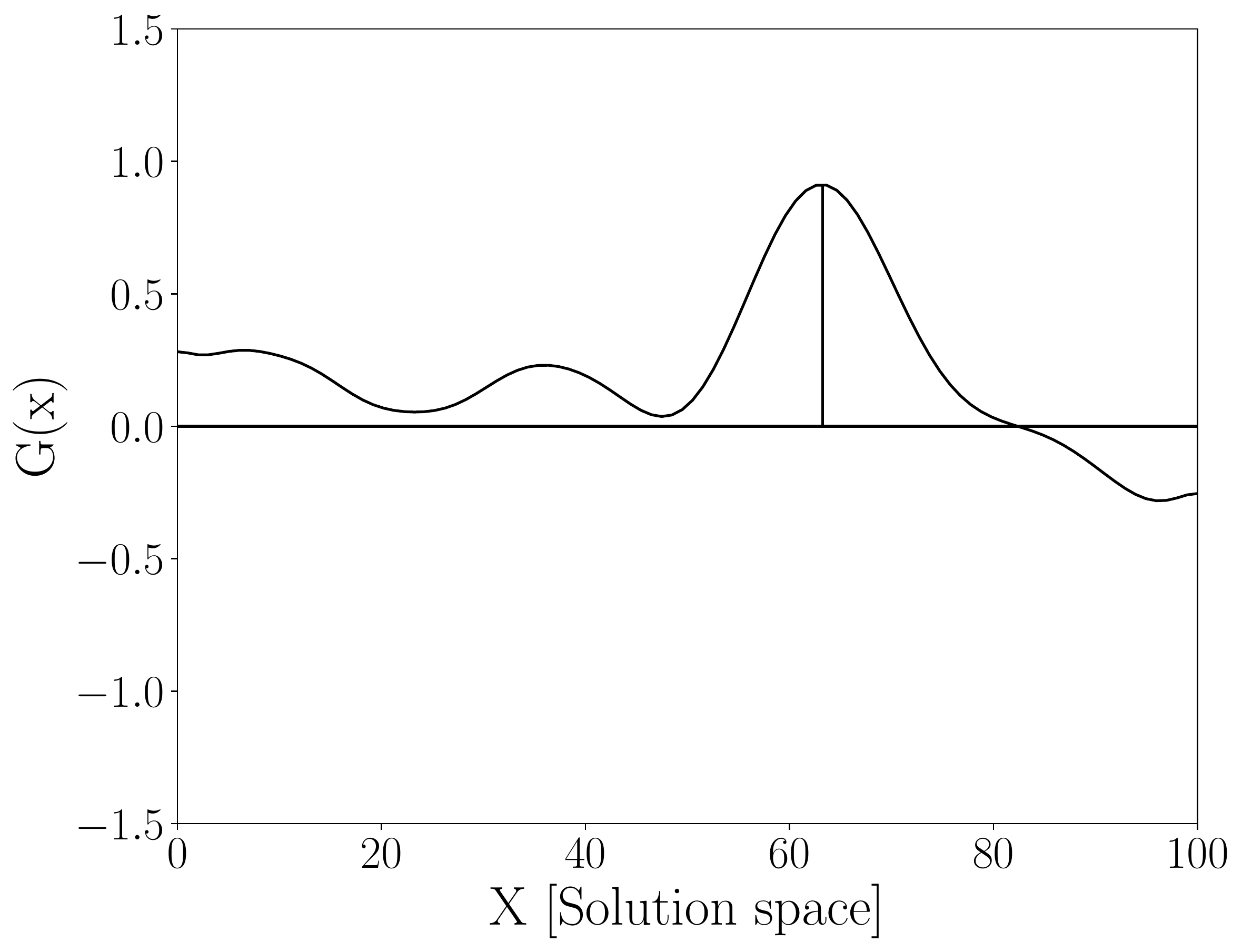}&
		\includegraphics[width=0.31\linewidth]{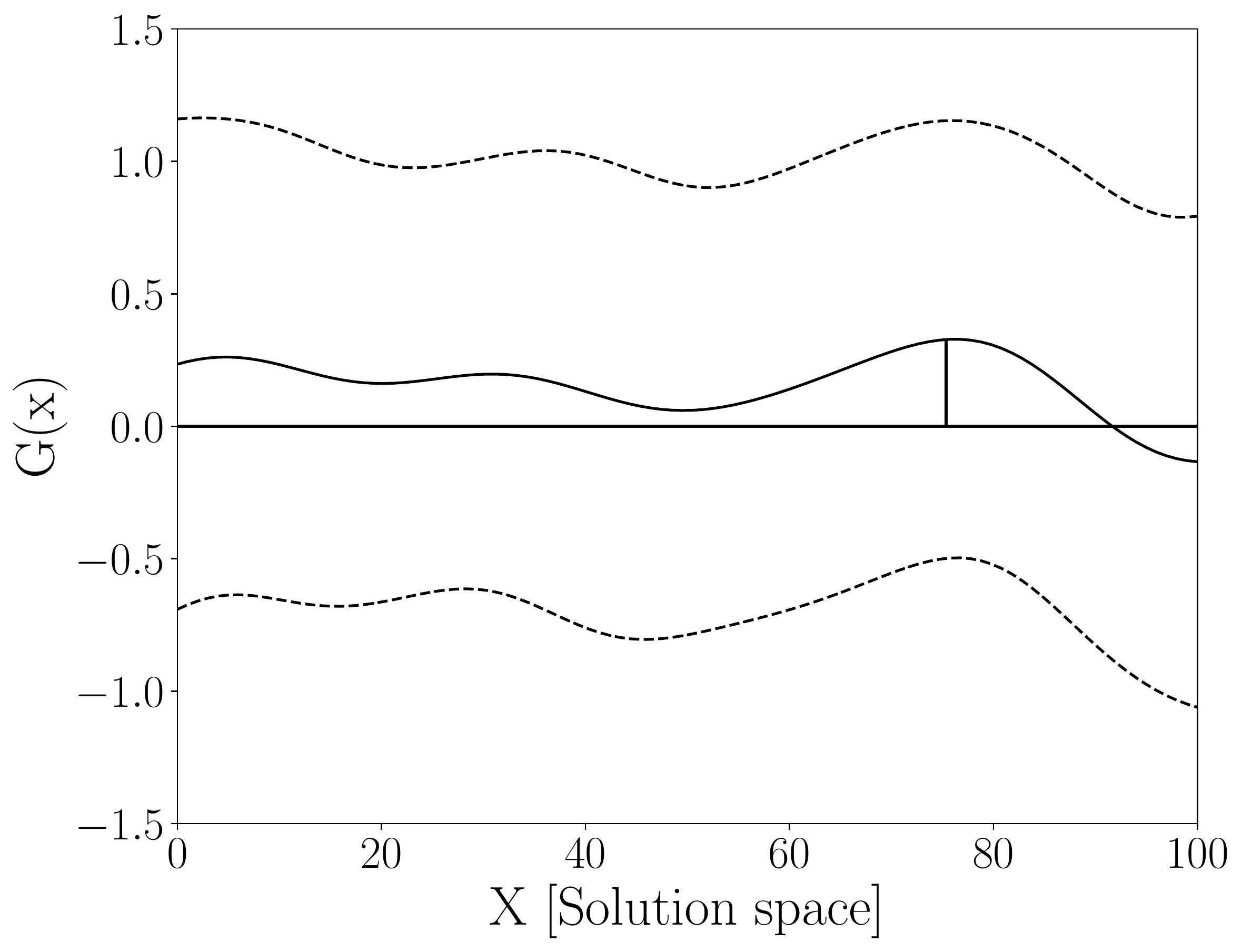}&
		\includegraphics[width=0.31\linewidth]{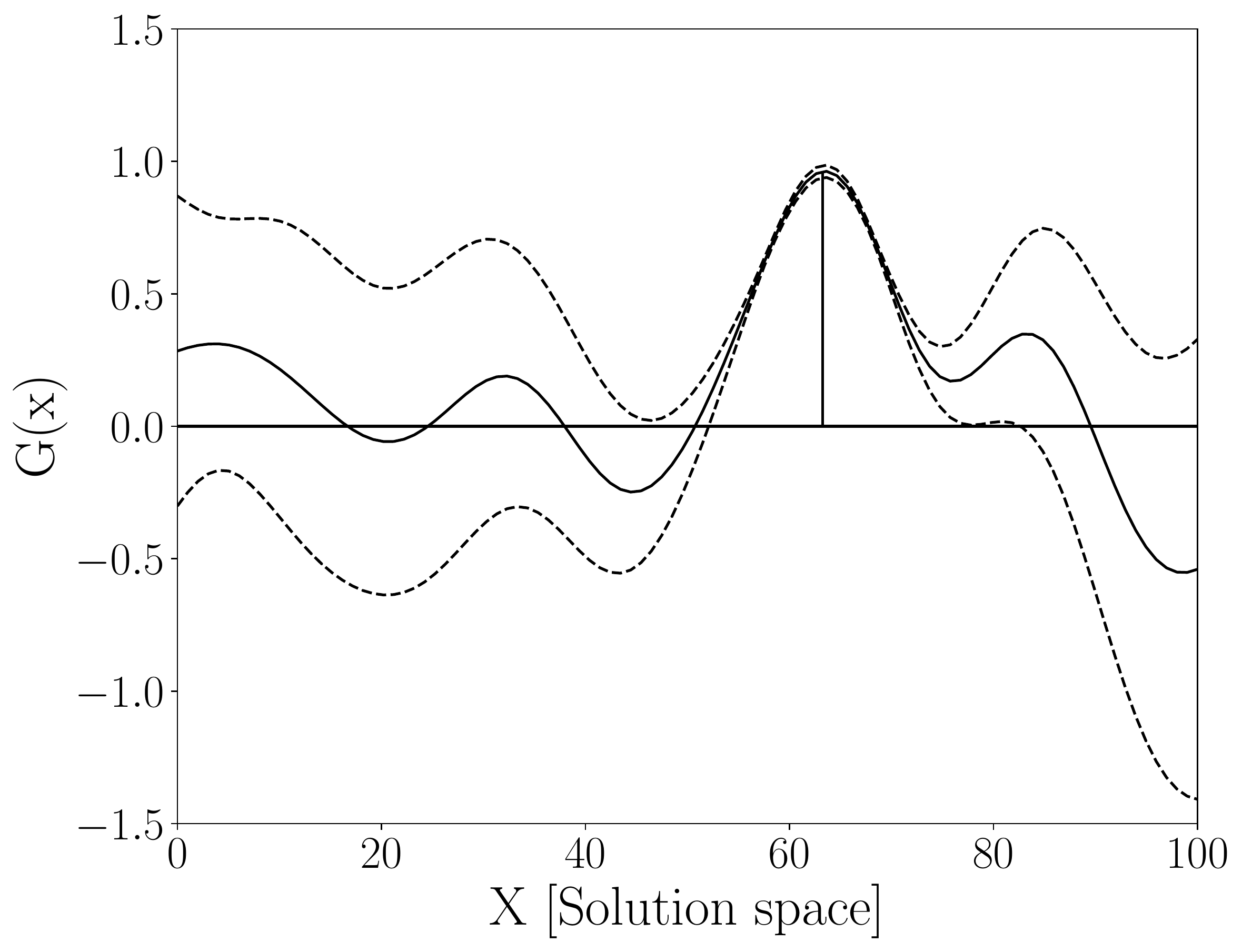}\\
		(d)&(e)&(f)\\
	\end{tabular}
	\captionof{figure}{In all plots, small points represent function evaluations. (a) \T $\theta(\bx,\ba)$, (d) shows $G(x)$ using the \T $\theta(\bx,\ba)$ and uniform \Pa. After 10 initial samples, (b) shows the surface $\mu^{10}(x,a)$, (e) $G(x, \Rc^{0},\F^{10})$. After 90 samples allocated by Equation~\ref{VoI_KG}, (c) shows the surface given by $\mu^{100}(x,a)$, (f) shows $G^{100}(x,\Rc^{0},\F^{100})$.}
	\label{fig:VoI_fixed_inputs}
	
\end{figure}

Fig. \ref{fig:VoI_fixed_inputs} shows Knowledge Gradient with fixed input uncertainty. At the start of sampling, initial samples are allocated by Latin hypercube sampling, the Gaussian process prediction of $\theta(\bx,\ba)$ and $G(\bx; \Rc^{0}, \F^{10})$ after the initial allocation are shown in Fig. \ref{fig:VoI_fixed_inputs}.b and \ref{fig:VoI_fixed_inputs}.e  assuming a uniform distribution for $\mathbb{P}[\ba]$. Then a budget of \B samples is allocated sequentially according to Eq.~\ref{VoI_KG}. (Fig \ref{fig:VoI_fixed_inputs}.c). Once all samples have been allocated, based on the learned Gaussian process model, the design $\bx$ with the largest predicted performance, according to Eq.~\ref{x_r}, is recommended to the user (Fig. \ref{fig:VoI_fixed_inputs}.f).


\subsection{Value of Information of Data from External Sources}\label{sec:VoIInfo}

\begin{figure}[bth]
	\centering
	\begin{tabular}{cc}
	    \includegraphics[width=0.48\linewidth]{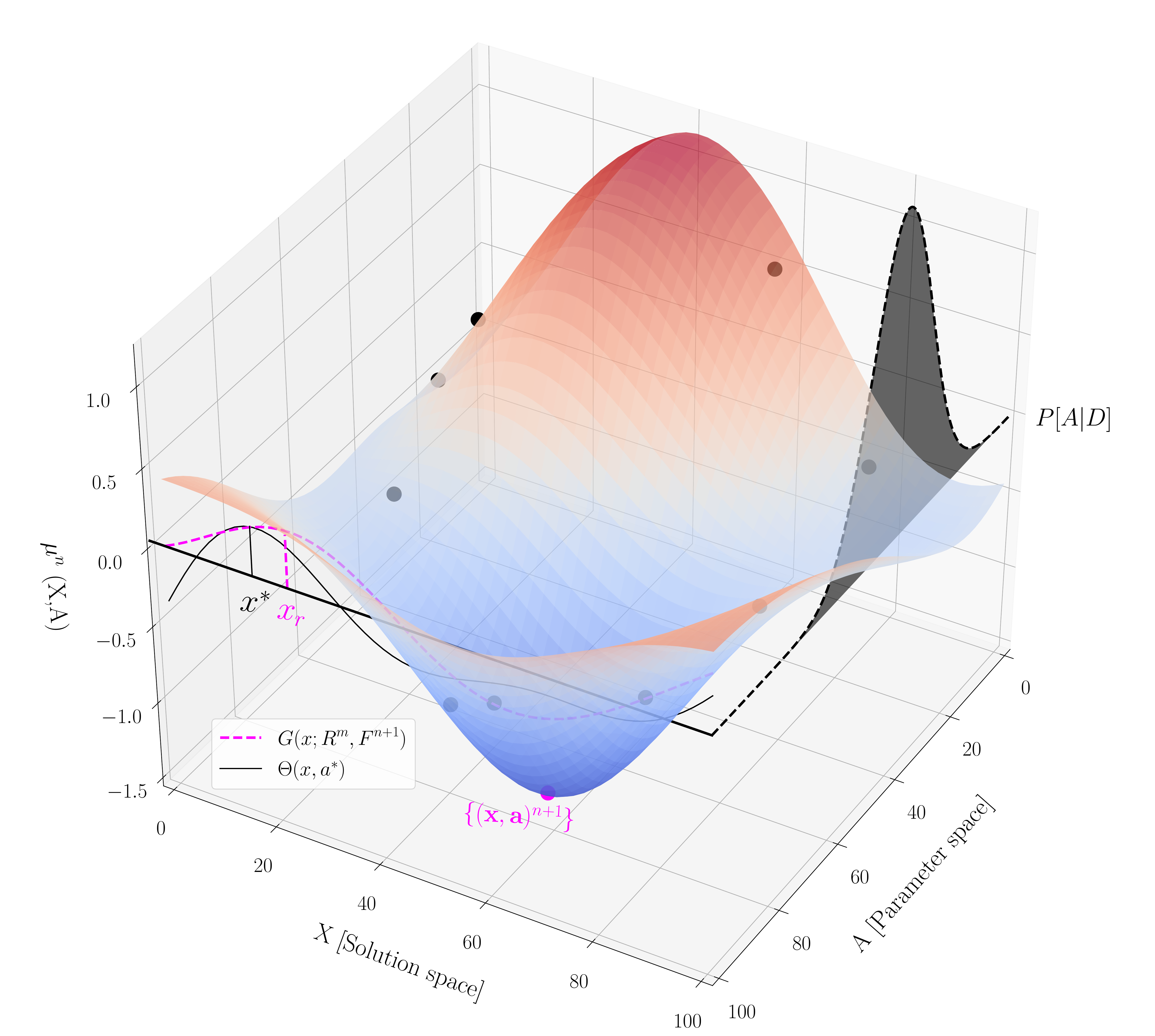}&
		\includegraphics[width=0.48\linewidth]{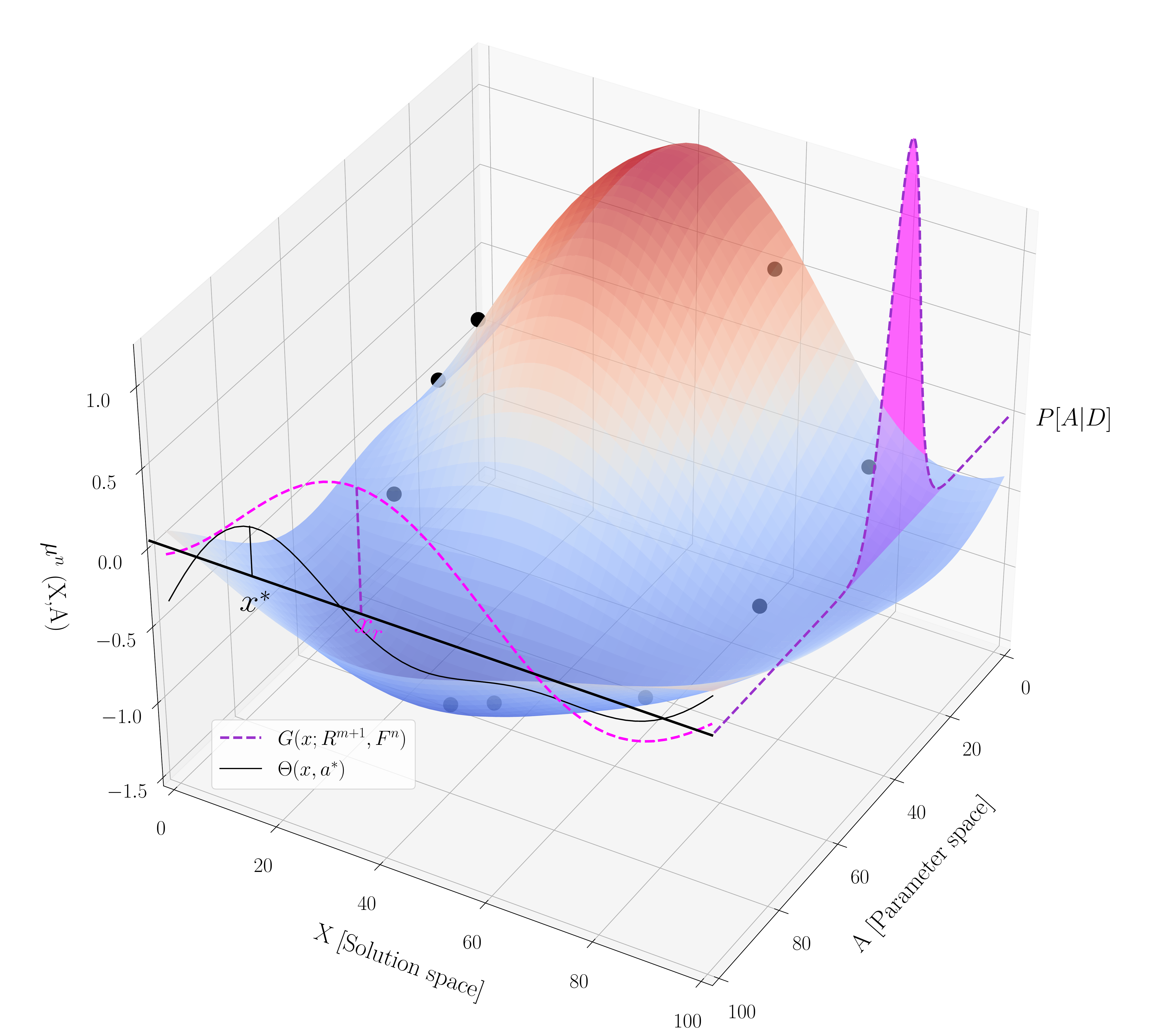}\\
		(a) & (b)\\
	\end{tabular}
	\captionof{figure}{Magenta shows the action chosen by the algorithm and how the model is updated. (a) A simulation run at $(x,a)^{n+1}$ is collected to update the Gaussian Process while leaving the current \Pa fixed. (b) A sample is collected from a \s while leaving the Gaussian process fixed, reducing just the \a uncertainty.
	\label{SampInputUncert}}
\end{figure}

Instead of collecting simulation data, we may collect data from a \s $r^{m+1}\sim \P[r|s^{m+1},\ba^*]$ thereby augmenting the corresponding dataset $\Rc^{m+1} = \Rm\cup \{(s,r)^{m+1}\}$. This also produces a non-negative improvement we denote $\VoI(s; \Rm, \Fn)$ leading to an increase in predicted performance. Figure \ref{SampInputUncert} shows the impact of both decisions. We refer to $\VoI(s; \Rm, \Fn)$ as the Value of Information of collecting additional external data,

\begin{eqnarray}
\VoI(s; \Rm, \Fn) &=& \E_{r^{m+1}}\left[\frac{ G^*(\Rc^{m+1}, \Fn)- G^*(\Rm, \Fn)}{c_{s}} \big| s, \Rm \right] 
\end{eqnarray}\label{VoI_data_source}
with $c_s$ being the cost of sampling external data source $s$. 

\begin{figure}[ht]
	\centering
	\begin{tabular}{c}
		\includegraphics[height=7.6cm, trim={100 80 150 100},clip]{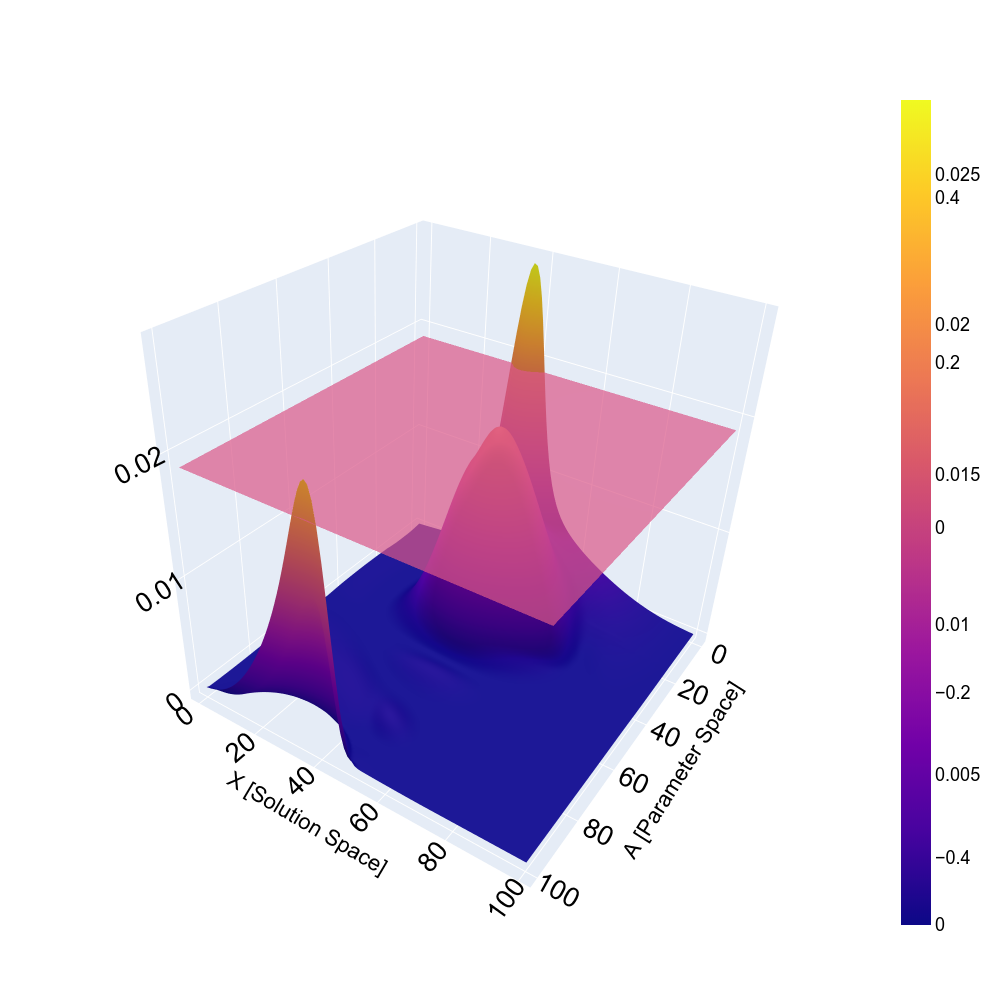}
	\end{tabular}
	\captionof{figure}{\ The multi-model function shows the Value of Information of simulation, $\VoI((x,a)^{n+1};\Rm,\Fn)$. A simulation point at $(100, 100)$ would yield little benefit while a simulation point at $(30, 0)$ would be very useful. The Value of Information for the \s, $\VoI(s;\Rm,\Fn)$ is computed separately and we overlay it here as the constant magenta plane. As the largest VoI for simulation, $\VoI^t((30, 0)) = 0.030$ is higher than for data source $\VoI^t(s)=0.019$, thus, for this iteration, the algorithm chooses to evaluate $f(30, 0)$. \label{DeltaLoss_figure}}
\end{figure}

In Section \ref{sec:VoISim}, the input parameter distribution is assumed to be fixed. However, better estimates of the input distribution to infer $a^*$ would yield  a final recommended solution closer to the true best solution. Figure \ref{SampInputUncert} shows that acquiring a new \sr $(s,r)^{m+1}$ changes the future \xr. The difference between $G^*(\Rm, \Fn)$ and the estimated realisations of $G^*(\Rc^{m+1}, \Fn)$ gives a non-negative difference that can be used to assess the benefit of sampling a \s $s$ against acquiring a \xa, as shown in Figure.\ref{DeltaLoss_figure}.

$\VoI(s;\Rm,\Fn)$ is computed using Monte-Carlo where samples $r^{m+1}$ are generated according to the predictive density $\P[r^{m+1}|s;\Rc^{m}]$ where each sample results in a new $G^{*}(x;\Rc^{m+1},\Fn)$. For the rest of this work we will use the shorthand $\VoI^t(\cdot) = \VoI(\cdot; \Rm, \Fn)$ to refer to the value of information at iteration $m+n=t$. We note that extending the method to account for multiple \ss is simply a case of computing $\VoI^t(\cdot)$ for each individual \s $s \in S$.

\subsection{Algorithm}\label{sec:Algorithm}

BICO is outlined in Algorithm \ref{alg:Bico}. On Line 1, the algorithm begins by fitting a Gaussian process model to a set of initial \xas $\Fn$ specified by a ‘space-filling’ experimental design, more specifically, we chose the initial set of \xas by a Latin hypercube sampling (LHS) design. Also, we compute the posterior \a distribution for any collected \s points $\Rm$. 

After initialisation, the algorithm continues in an optimisation loop until all the budget $\B$ has been consumed. During each iteration, we compute the \VoI ~of collecting a new \xa $(\bx^{n+1},\ba^{n+1},y^{n+1})$ according to $\VoI^{t}((\bx,\ba))$ (Line 2) and the \VoI ~of
collecting  a new sample for each one of the \ss $s \in S$ $\VoI^{t}(s)$ (Line 3). The action that gives greater value determines whether we collect a sample $(\bx,\ba,y)^{n+1}$ or $r^{m+1}$. In the first case, the Gaussian process model is updated according to the new \x sample (Lines 4-6) and, for the second case, the posterior \a distribution is updated according to the new \s sample (Lines 7-9). At the end of \B samples, the design $\bx$ with the largest predicted performance $G(\bx)$ is recommended to the user (Line 10).

\begin{algorithm}
	\caption{BICO algorithm}\label{alg:Bico}

	\KwIn{action space: $X\times A$, $s_1,...,s_{N_s}$, actions costs $\{c_{f},c_{1},\dots,c_{N_{s}}\}$, budget $\B$, initial data $\Rm$, $\Fn$}
	\vspace{1.2mm}
    1. Fit a Gaussian process to $\Fn$ and compute a posterior distribution $\P[\ba|\Rm]$.\\
    \vspace{1.2mm}
	\textbf{While} \text{$b$ < \B}:\\
	\vspace{1.0mm}
    \hspace{5mm}2. Compute $(\bx,\ba)^{n+1} = \arg \max_{(\bx,\ba) \in X \times A}\VoI^{t}((\bx,\ba))$.\\
    \vspace{0.5mm}
    \hspace{5mm}3. Compute $s^{m+1} = \arg \max_{s \in S}\VoI^{t}(s)$
    
    \vspace{1.2mm}
    \hspace{5mm}\textbf{If }
    $\max_{(\bx,\ba) \in X \times A}\VoI^{t}((\bx,\ba))> \max_{s \in S}\VoI^{t}(s)$\\
    \vspace{1.0mm}
    \hspace{7mm}\text{4. Collect from simulator }$(\bx,\ba, y)^{n+1}$ \\
    \vspace{0.5mm}
    \hspace{7mm}\text{5. } $\mathscr{F}^{n+1} = \mathscr{F}^{n} \cup \{(\bx,\ba,y)^{n+1}\}$\\
    \vspace{0.5mm}
    \hspace{7mm}\text{6.  Fit a Gaussian process to $\mathscr{F}^{n+1}$} \\
    \vspace{0.5mm}
    \hspace{7mm}\text{7. }Update budget consumed $b \gets b + c_{f}, n \gets n+1$\\
    
    \vspace{1.2mm}
    \hspace{5mm}\textbf{Else }\\
    \vspace{1.0mm}
    \hspace{7mm}\text{8. Collect from \s }$(s, r)^{m+1}$ \\
    \vspace{0.5mm}
    \hspace{7mm}\text{9.} $\mathscr{R}^{m+1} = \mathscr{R}^{m} \cup \{(s,r)^{m+1}\}$\\
    \vspace{0.5mm}
    \hspace{7mm}\text{10. }Compute a posterior distribution $\P[\ba|\Rc^{m+1}]$\\
    \vspace{0.5mm}
    \hspace{7mm}\text{11. }Update budget consumed $b \gets b + c_{s^{m+1}}, m \gets m+1$\\
    \vspace{1.2mm}
 10. Recommend $\bx_{r} = \arg \max_{x} G(\bx; \mathscr{R}^{m},\mathscr{F}^{n})$

\end{algorithm}

\subsection{Properties of BICO} \label{sec:BICOproofs}

In the Appendix we proof consistency of BICO, however we outline the main findings here. We specifically show that if $X$ is discrete and $A \subset \mathbb{R}^{d}$ is continuous, the BICO algorithm will find the true optimal solution $x^*$ as well as the true parameters $a^*$.  This build on a previous proof by \citeauthor{toscanopalmerin2018bayesian} (\citeyear{toscanopalmerin2018bayesian}) that shows consistency for input uncertainty and collection of \xasNospace.

Proposition \ref{conv_s} shows that if a single action is performed infinitely often, then the value of performing the action vanishes $\lim_{t\to \infty}\VoI(\cdot; \Rm,\Fn)\to 0$. This implies the value of all actions eventually vanishes.\\

\textbf{Proposition \ref{conv_s}.}
Let $a' \in A$, $x' \in X$ and $s \in \{1,\dots,N_{s}\}$ and suppose that $(x',a')$ or $s$ is observed infinitely often. Then $\VoI^{t}(\cdot;\Rm, \Fn) \rightarrow 0$ as $t \rightarrow \infty$.\\

Furthermore, if the Value of Information of all actions is zero, this implies that $x^*$ and $a^{*}$ is known.\\

\textbf{Proposition \ref{reverese_xstar}.} 
If $\lim_{n \rightarrow \infty} \VoI^{t}((x,a);\Rm,\Fn)= 0$ and $\lim_{m \rightarrow \infty} \VoI^{t}(s;\Rm,\Fn)= 0$ for all $(x,a)$ and $s$, then
$argmax_{x \in X}G(x;\Rm,\F^{\infty}) =argmax_{x \in X}\int_{A}\theta(x,a)\P[a|\Rm]da$ and $a^{*}$ is known.\\

Additionally, if we use a square exponential kernel, then the hyperparameters determine the relevance of \ss. Therefore,  non-relevant \ss will not be sampled from BICO.\\

\textbf{Remark \ref{noninfluentiala}.} 
Assuming a squared exponential kernel,

\begin{align}
\begin{split}
k^{0}((x,a)(x,a)')&=\sigma^{2}_{f}e^{-\frac{1}{2}\big (\frac{(x-x')^{2}}{l_{x}}+\frac{(a-a')^{2}}{l_{a}} \big ) }
\end{split}
\end{align}

and without loss of generality a \a $a \in A$, and a \x $x \in X$. Then $\VoI^t(s, \Rm,\Fn) = 0$ as $l_{a} \rightarrow \infty$.

  	\section{RESULTS AND DISCUSSION}\label{sec:NUMERICAL}

To demonstrate the performance of our BICO algorithm, we compare it against first collecting a percentage $p$ of the total budget to sample and update the input posterior distribution before the simulation optimisation begins. Then, the remaining budget $\B(1-p)$ is dedicated to sequentially sample from the \fNospace. For two or more input distributions, the initial portion, $\B p$, is evenly distributed over the different inputs, i.e., if we take an initial sample size of 30 data points to update three \asNospace, then each \a would be updated with 10 \rsNospace. For all experiments we consider 100 replications for the BICO algorithm and benchmark method.
Note that it is generally not possible to know in advance which proportion of the available budget should be allocated to external data collection, so different values of $p$ need to be tested. 

\subsection{GP-Generated Experiments}

To test Knowledge Gradient with fixed input uncertainty, we consider a test function with \x space $X = [0,100]$ and either one \a in $A = [0,100]$ or two parameters with $A = [0,100]^2$
generated from a Gaussian process with a squared exponential kernel with known hyper-parameters $l_{XA} = 10$, $\sigma^2_{0}=1$ , $\sigma^2_{\epsilon} = (0.1)^2$. The total budget in both cases was set to $\B=100$, and the cost to query a simulation or data source is assumed to be identical and equal to 1.
To model input uncertainty, we assume a uniform prior $\mathbb{P}[a] = \frac{1}{100}$ and normally distributed \rs for each \sNospace.

\begin{figure}
	\centering
	\begin{tabular}{cc}
	\includegraphics[width=0.45\linewidth]{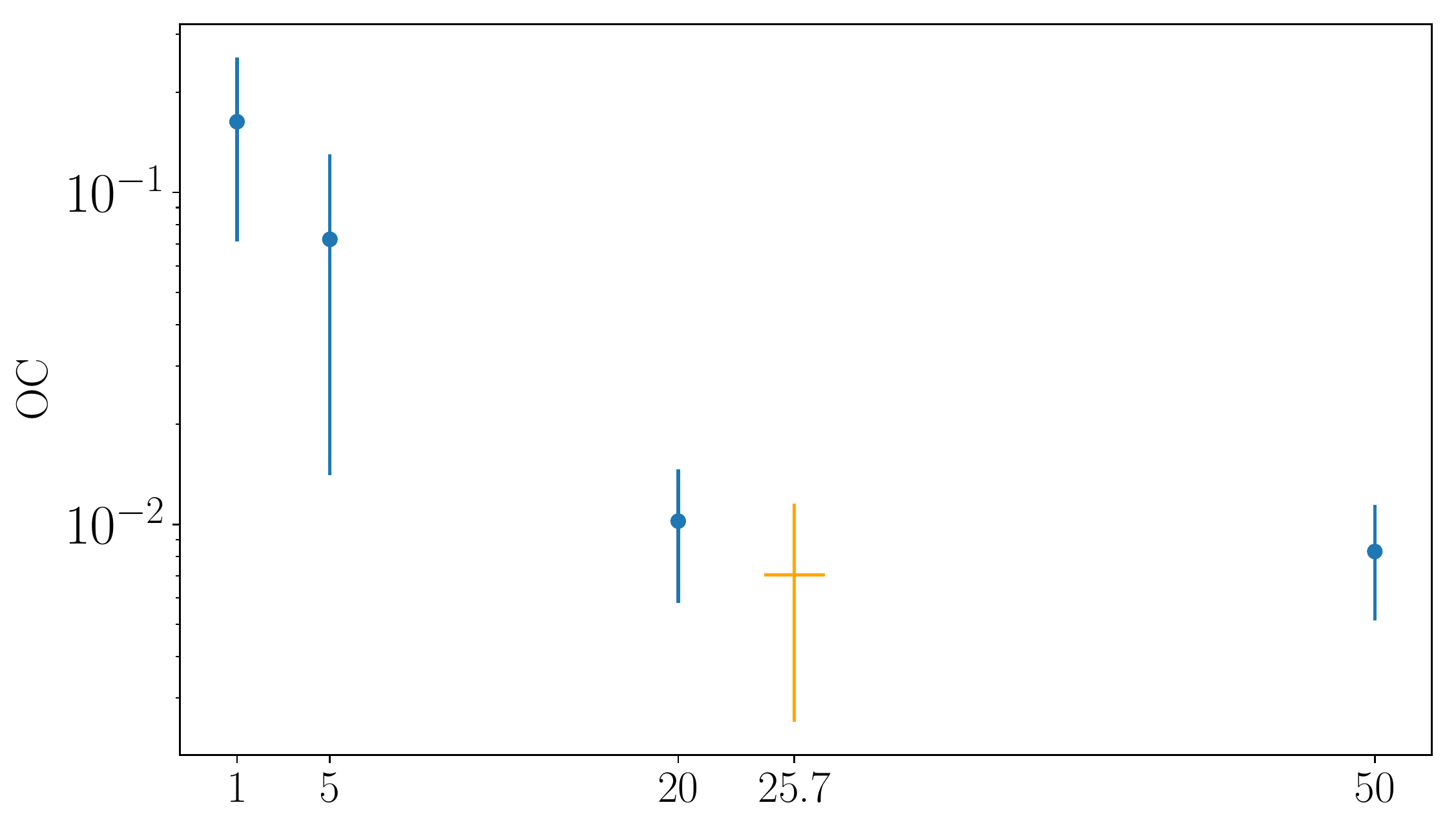}&
	\includegraphics[width=0.47
	\linewidth]{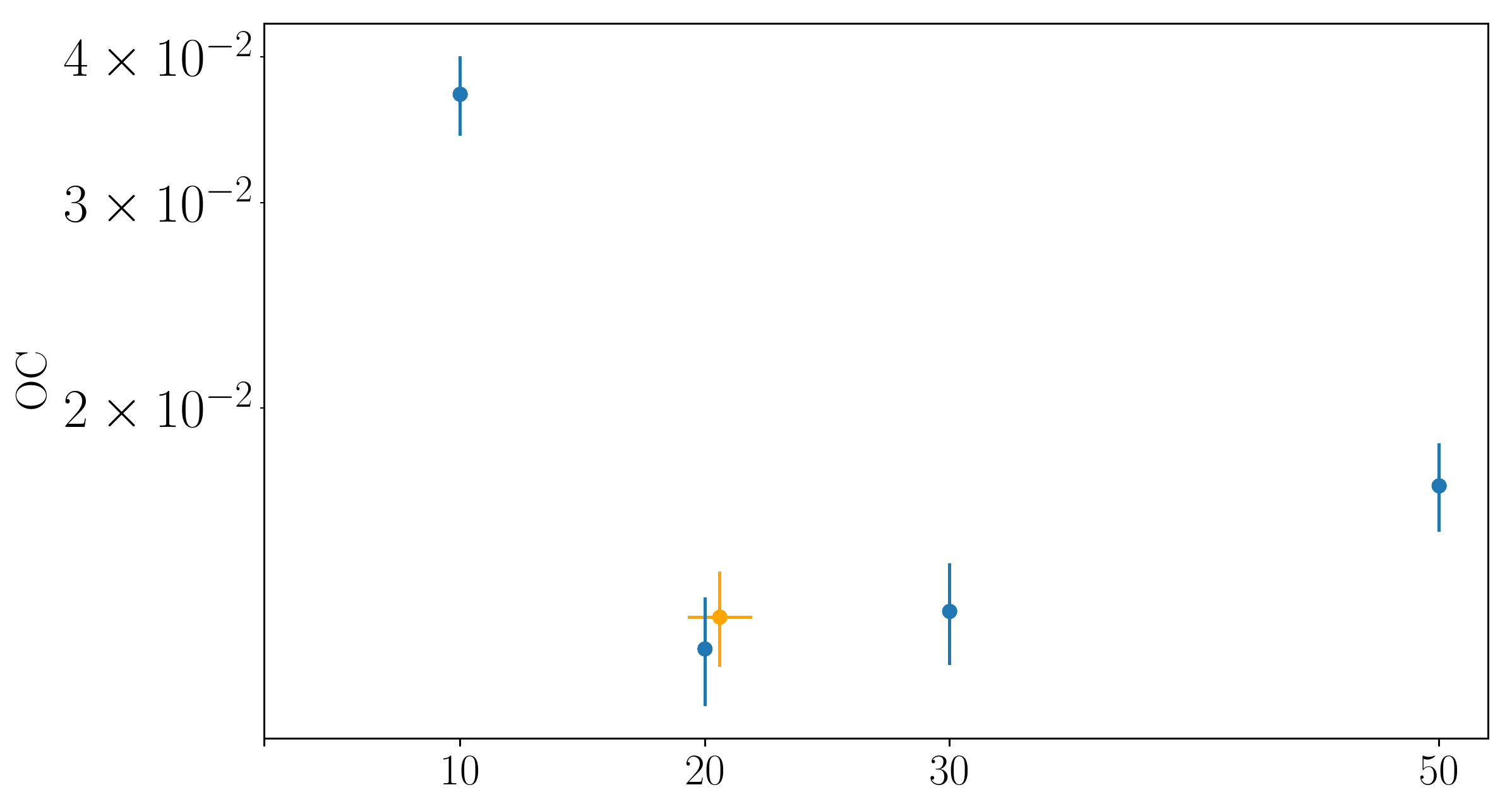}\\
	(a)&(b)
		
	\end{tabular}
	\captionof{figure}{Mean and 95\% CI for the OC plotted in a semilog scale for $\B=100$. (a) GP-Generated experiment with one \x and one \aNospace. (b) GP-Generated experiment with one \x and two \asNospace. In each experiment, each \a has a \sNospace.}
	\label{Results2}
\end{figure}

Results are shown in logarithmic scale in Figure \ref{Results2}, on the left for the case of a single \aNospace, on the right the case of two \asNospace. The horizontal axis shows the number of samples $m$ allocated to the \s to update $\P[a|\Rm]$, whereas the vertical axis shows the confidence interval of the OC after the budget $\B$ has been completely allocated. In both cases, BICO balances the sampling allocation effort in a sensible way, finding comparable results to taking the optimal initial number of samples $Bp$ to sample the \ssNospace. Somewhat surprisingly, it seems more effort should be allocated to data collection if there is only one data source. This is probably because in case of two data sources, the space over which the \f is defined is higher, requiring more effort to build a credible Gaussian process model.

\subsection{Newsvendor Simulation Optimisation}

Here, we consider the problem of a newspaper vendor, or any product that loses value very fast, who must decide how many copies of the day's paper to stock in the face of uncertain demand where any unsold copies will be worthless at the end of the day. If the \x $x$ is the number of newspapers ordered and a random demand $C \sim N(\mu,\sigma^{2})$, then  the profit $f(x,C)$ is given as,

$$
f(x,C) = p \min(x,C) - lx
$$

where, $p$ is the price and $l$ the production/purchase cost of a newspaper, with $p>l$. For this experiment we set  $p=$ 5, $l =$ 3 and $x \in [0, 100]$. We considered an initial allocation of 10 samples to train the Gaussian process model from an overall budget of $\B=50$ with uncertain mean $\mu$, with true value $\mu^{*}$=40 and known variance $\sigma^{2}$=10. In contrast with the previous experiment, results for BICO (orange) also show the average number of samples and its error bar as horizontal lines. Also in this experiment, BICO (orange) manages to allocate the budget $\B$ close to an adequate fixed initial number of samples (blue). 

\begin{figure}
	\centering
	\begin{tabular}{c}
	\includegraphics[width=0.66
	\linewidth]{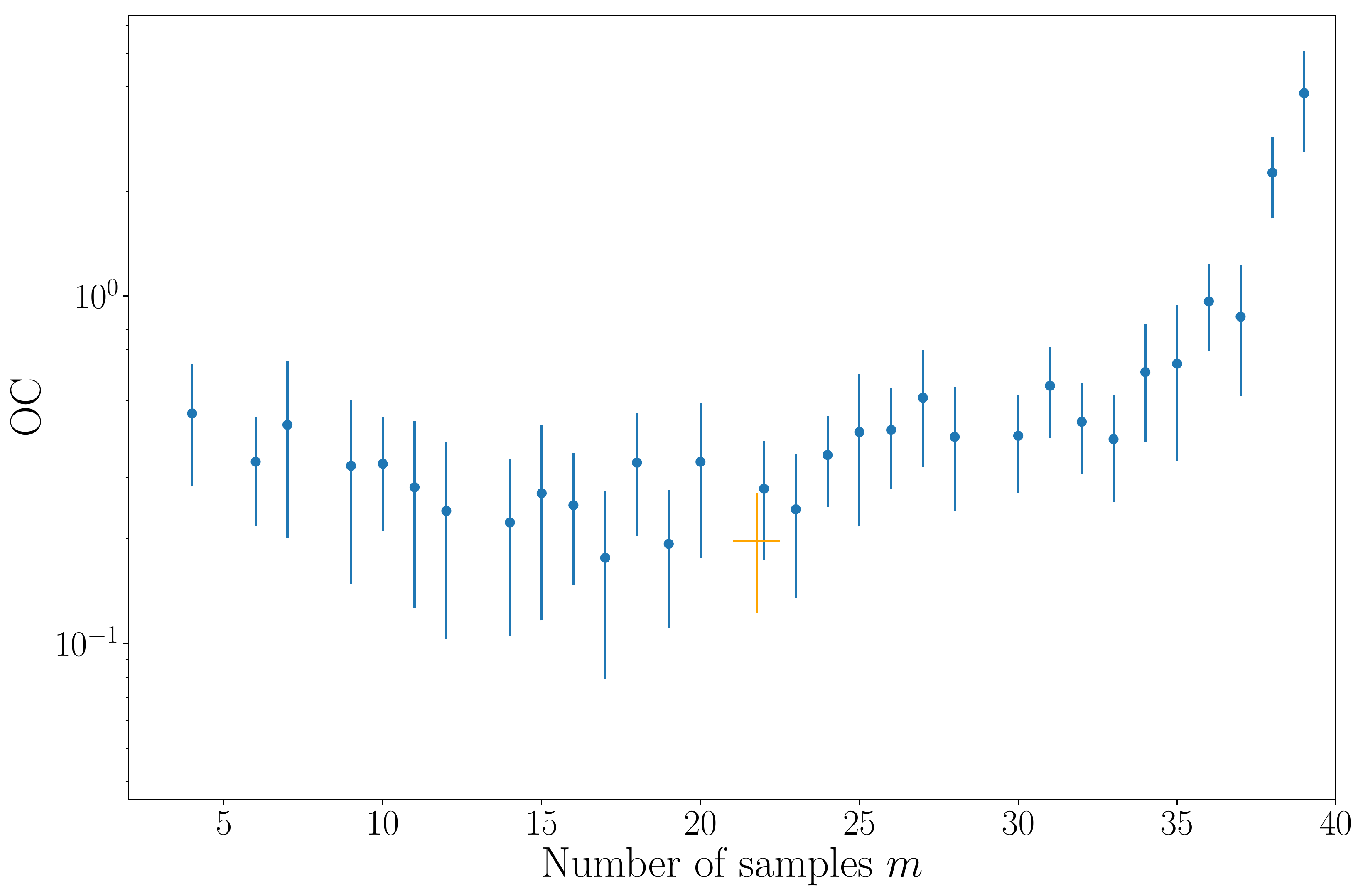}\\
		
	\end{tabular}
	\captionof{figure}{Mean and 95\% CI of OC where $B=50$. Blue points: at the start, $m$ input parameter samples are collected to estimate $a^*$  and thereafter standard Knowledge Gradient for (fixed) input uncertainty is applied to allocate the remaining budget to simulation points. There is no automatic trade-off between data types, $m$ must be user specified. Orange: BICO algorithm, the horizontal confidence interval showing the range of sample sizes $m$ chosen by BICO. BICO automatically avoids allocating too much budget to input samples, consistently avoiding the inferior range $m>30$.}
	\label{RESULTS_newsvendor}
\end{figure}

 	\section{CONCLUSION}\label{sec:CONCLUSION}

In this paper, we proposed a novel unified simulation optimisation algorithm that, in each iteration, automatically determines whether to perform more simulation experiments or instead collect more real world data to reduce the uncertainty about the input parameters. A comparison with
an algorithm that allocates a fixed, pre-determined fraction $p$ of the available budget to external data collection 
demonstrated that BICO's allocation mechanism is very powerful and results in a solution performance and fraction of budget allocated to external data collection similar to what can be achieved with the optimal allocation, which is generally not know in advance.


There are some interesting extensions of this work with concrete practical applications which are possible to pursue. 
One example is the extension to multi-objective optimisation, where the uncertainty
about a user's preferences over objectives can be reduced by querying the user.
While we assumed in this paper that the design space and
the input distribution parameter space can each be described by continuous parameters, the proposed methods should also be tested with discrete parameters.

 	\section*{Acknowledgements} 
  The first author would like to acknowledge funding from EPSRC through grant EP/L015374/1.

    \bibliographystyle{ACM-Reference-Format}
  
    \bibliography{bibliography}
        
     \appendix
\section{Appendix}\label{Appendix}

\subsection{BICO Convergence}

In this section, we show consistency of BICO. We specifically show that if $X$ is discrete and $A \subset \mathbb{R}^{d}$ is continuous, then when given an infinite sampling budget $B$, or $t\to \infty$, the BICO algorithm will find the true optimal solution $x^*$ as well as the true parameters $a^*$. 

The proof is composed of three parts, firstly, Remark~\ref{thm:VOI_positive} shows that the $\VoI(\cdot; \Rm,\Fn)$ for any action is non-negative which follows naturally from Jensen’s inequality. Second in part we show that if a single action is performed infinitely often, then the value of performing the action vanishes $\lim_{t\to \infty}\VoI(\cdot; \Rm,\Fn)\to 0$. Together these imply that any action repeated infinitely often results in that action becoming a \emph{minimum} of the $\VoI(\cdot)$ function and since BICO performs the action that is a \emph{maximum} of $\VoI(\cdot)$, the value of all actions eventually vanishes. Thirdly and finally, if the value of all actions is zero, this implies that $x^*$ is known.

The first remark shows that the $\VoI(s; \Rm,\Fn)$ is non-negative, meaning that there is always a benefit in collecting more data

\begin{ppr}\label{thm:VOI_positive}
$\VoI^{t}(\cdot; \Rm, \Fn) \geq 0$, for $s \in \{1,\dots,N_{s}\}$ and $(x,a) \in X \times A$\label{VoIpost}
\end{ppr}

\vspace{1 mm}
\textit{Proof of Remark \ref{VoIpost}}

The proof for both types of action follows from the tower property and Jensen's inequality. Using the Tower property and $\E_{y^{n+1}}[\mu^{n+1}(x,a)] = \mu^{n}(x,a)$,
we first prove the result for simulation data $\VoI^t((x,a))\geq 0$

\begin{align}
\begin{split}
\E_{y^{n+1}}\left[ \max_{x}G(x;\Rm, \F^{n+1}) \big| (x,a), \Fn \right] &\geq \max_{x}\E_{y^{n+1}}\left[G(x;\Rm, \F^{n+1}) \big| (x,a), \Fn \right]
\end{split}\\
\begin{split}
    &=\max_{x}\E_{y^{n+1}}[\E_{a}[\mu^{n+1}(x,a)]]
\end{split}\\
\begin{split}
    &=\max_{x}\E_{a}[\E_{y^{n+1}}[\mu^{n+1}(x,a)]]
\end{split}\\
\begin{split}
    &=\max_{x}\E_{a}[\mu^{n}(x,a)]
\end{split}\\
\begin{split}
    &=\max_{x}G(x;\Fn,\Rm)
\end{split}
\end{align}

For $\VoI^{t}(s)\geq 0$,

\begin{align}
\begin{split}
\E_{r^{m+1}}\left[ \max_{x}G(x;\Rc^{m+1}, \Fn) \big| s, \Rm \right] &\geq \max_{x}\E_{r^{m+1}}\left[G(x;\Rc^{m+1}, \Fn) \big| s, \Rm \right]
\end{split}\\
\begin{split}
    &=\max_{x}\int_{r^{m+1}}\int_{A}\mu^{n}(\bx,\ba)\P[\ba|\Rc^{m+1}]\P[r^{m+1}|\Rm]d\ba dr^{m+1}
\end{split}\\
\begin{split}
    &=\max_{x}\int_{A}\mu^{n}(\bx,\ba)\int_{r^{m+1}}\P[\ba|\Rc^{m+1}]\P[r^{m+1}|\Rm]dr^{m+1} d\ba 
\end{split}\\
\begin{split}
    &=\max_{x}\int_{A}\mu^{n}(\bx,\ba)\P[\ba|\Rm]d\ba 
\end{split}\\
\begin{split}
    &=\max_{x}G(x;\Rm, \Fn)
\end{split}\qed
\end{align}

The second part of the convergence proof shows that if an action is performed infinitely often, the value of performing the action tends to zero, the information gain in repeating an action decreases and eventually becomes a minimum of the $VoI^t(\cdot)$ function. This is shown in two stages, firstly that $\P[a|\Rm]\to \delta_{a^{*}=a}$ which then implies $\VoI^t(s)=0$ and secondly, sampling $(x,a)'$ infinitely often implies $\VoI^t((x,a))=0$

In order to prove the proposition, we rely on  Theorem \ref{consistencty} and Proposition \ref{existence_limit}. Theorem \ref{consistencty} states that the \a distribution $\P[a|\Rm]$ converges to $\delta_{a=a^{*}}$ as $m$ increases, and Proposition. \ref{existence_limit} establishes the limit of $\mu^{n}(x,a)$ and $\tilde{\Sigma}^{n}(x;(x,a))$ as $n \rightarrow \infty$

\begin{thm}
If $\ba$ is defined on a compact set and $C$ is a neighbourhood of $\ba^{*}$ with nonzero prior probability, then $\P(a \in C | \Rm) \rightarrow 1$ as $m \rightarrow \infty$, where $\ba^{*}$ is the value of $C$ that minimises the KL divergence. \label{consistencty}
\end{thm}
A proof of consistency of the posterior distribution is a standard result (see Appendix B in \citeauthor{BayesDataAnal}\citeyear{BayesDataAnal}) and is omitted for brevity.

\begin{ppr}
Let $x,x' \in X$; $a,a' \in A$, and $n\in \mathbb{N}$. The limits of the series $(\mu^{n}(x,a))$ and $(V^{n}((x,a),(x',a')))$ (shown below) exist.

\begin{align}
    \begin{split}
        \mu^{n}(x,a) &= \E_{n}[f(x,a)]
    \end{split}
\end{align}

\begin{align}
    \begin{split}
        V^{n}((x,a),(x',a')) &=\E_{n}[f(x,a)\cdot f(x',a')]
    \end{split}\\
    \begin{split}
        &= k^{n}((x,a),(x',a'))+ \mu^{n}(x,a)\cdot \mu^{n}(x',a') 
    \end{split}
\end{align}

Denote their limits by $\mu^{\infty}(x,a)$ and $V^{\infty}=((x,a),(x',a'))$ respectively. 

\begin{align}
    \begin{split}
        \lim_{n \rightarrow \infty} \mu^{n}(x,a)&=\mu^{\infty}(x,a)
    \end{split}\\
    \begin{split}
        \lim_{n \rightarrow \infty} V^{n}((x,a),(x',a'))&=V^{\infty}((x,a),(x',a'))
    \end{split}
\end{align}

If $(x',a')$ is sampled infinitely often, then $\lim_{n\rightarrow \infty} V^{n}((x,a),(x',a')) = \mu^{\infty}(x,a)\cdot 
\mu^{\infty}(x',a')$ holds almost surely.

\label{existence_limit} 
\end{ppr}

\textit{Proof} 

\citeauthor{StochProc}(\citeyear{StochProc}) states in Proposition 2.8 that any sequence of conditional expectations of an
integrable random variable under an increasing  convex function is a uniformly integrable martingale. Thus,
both sequences converge almost surely to their respective limit. If $(x', a')$ is sampled infinitely often, then its posterior variance goes to zero,
and $\E_{n}\big[f(x,a)\cdot f(x',a') \big] \rightarrow \mu^{\infty}(x,a) \cdot \mu^{\infty}(x',a')$.

The following propositions show consistency of BICO. More specifically, Proposition \ref{conv_s} shows that observing either $(x',a')$ or $s$ infinitely often will make $VoI^{t}(\cdot)$ converge to zero for that specific action. Proposition \ref{reverese_xstar} shows that if $VoI^{t}(\cdot)=0$ as $t \rightarrow \infty$ for any action, then the global optimiser $x^{*}$ and $a^{*}$ are known. All the results assume that the lenghtscale of the kernel function is bounded, $l_{x}<\infty$ and $l_{a}<\infty$.

\begin{ppr}
Let $a' \in A$, $x' \in X$ and $s \in \{1,\dots,N_{s}\}$ and suppose that $(x',a')$ or $s$ is observed infinitely often. Therefore $\VoI^{t}(\cdot;\Rm, \Fn) \rightarrow 0$ as $t \rightarrow \infty$.\label{conv_s}
\end{ppr}

\textit{Proof of Proposition} \ref{conv_s}:

We first prove the result when simulation data $(x,a)$ is infinitely sampled. Let's first consider the case where there is no noise in the simulation output, i.e, $\sigma_{\epsilon} = 0$. So, if $(x,a)'$ was previously observed and belongs to $\Fn$, then sampling on the same location $(x,a)^{n+1}=(x,a)'$ will not change the posterior variance.

\begin{align}
    \begin{split}
        \tilde{\Sigma}^{n}(x;(x,a)^{n+1})^{2}&= Var_{n}[G(x; \Rm, \F^{n+1}) | (x,a)^{n+1}]
    \end{split}\\
    \begin{split}
        &= Var_{n}[\E_{n+1}[\E_{a}[\theta(\bx,\ba)|\Dm]] | (x,a)^{n+1}]
    \end{split}\\
    \begin{split}
      &= Var_{n}[\E_{a}[\theta(\bx,\ba)|\Dm]] - \E_{n}[Var_{n+1}[\E_{a}[\theta(\bx,\ba)|\Dm]] | (x,a)^{n+1}]
    \end{split}\\
    \begin{split}
      &= Var_{n}[\E_{a}[\theta(\bx,\ba)|\Dm]] - Var_{n+1}[\E_{a}[\theta(\bx,\ba)|\Dm]| (x,a)^{n+1}]
    \end{split}\\
    \begin{split}
      &= \int\int k^n((x,a);(x,a)'')dada'' - \int\int k^{n+1}((x,a);(x,a)'')dada''
    \end{split}
\end{align}

It follows that $\tilde{\Sigma}^{n}(\bx;(x,a)^{n+1})=0$ because $k^n((x,a);(x,a)'') = k^{n+1}((x,a);(x,a)'')$ when $(x,a)^{n+1} = (x,a)'$. Therefore, $G(x;(x,a)^{n+1})=G(x;(x,a)^{n})$ for any sample $(x',a',y^{n+1})$ producing $VoI((x,a)',\Rm,\Fn)=0$. Now let's assume $\sigma_{\epsilon} > 0$ and $(x,a)'$ was observed infinitely often, 

\begin{align}
    \begin{split}
        \lim_{n  \rightarrow \infty}\tilde{\Sigma}^{n}(x;(x,a)') &= \lim_{n \rightarrow \infty}\int_{A}\tilde{\sigma}^{n}((x,a);(x,a)')\P[a|\Rm]da
    \end{split}
\end{align}

Since $\tilde{\sigma}^{n}((x,a);(x,a)')$ is a uniformly integrable (u.i.) random variable,
\begin{align}
    \begin{split}
       \lim_{n \rightarrow \infty}\int_{A}\tilde{\sigma}^{n}((x,a);(x,a)')\P[a|\Rm]da&= \int_{A}\lim_{n \rightarrow \infty}\tilde{\sigma}^{n}((x,a);(x,a)')\P[a|\Rm]da
    \end{split}\\
    \begin{split}
       &= \int_{A}\lim_{n \rightarrow \infty}\frac{k^{n}((x,a);(x,a)')}{\sqrt{k^{n}((x,a)';(x,a)') + \sigma^{2}_{\epsilon}}}\P[a|\Rm]da
    \end{split}\\
    \begin{split}
        &=0
    \end{split}
\end{align}

Considering that $\mu^{n}(x,a)$ and $\Sigma^{n}(x;(x,a)') $ are uniformly integrable (u.i.) families of random variables
that converge a.s. to their limits $\mu^{\infty}(x,a)$ and $\Sigma^{\infty}(x;(x,a)')=0 $, where $\lim_{n\rightarrow \infty}G(x;\Fn \Rm) = G(x;\F^{\infty} \Rm)$.

\begin{align}
    \begin{split}
        \lim_{n \rightarrow \infty} VoI((x,a);\Fn,\Rm) &= \frac{\int_{-\infty}^{\infty}\phi(Z)\max_{x''}\{G(x;\F^{\infty} \Rm)+\Sigma^{\infty}(x;(x,a)')Z\} - \max_{x''}\{G(x;\F^{\infty} \Rm)\}}{c_{f}}
    \end{split} 
\end{align}

Since $Z$ and $\Sigma^{\infty}(x;(x,a)')$ are both independent and u.i,  $\Sigma^{\infty}(x;(x,a)')Z$ is u.i, also the sum of u.i. random variables is u.i., and the maximum over a finite collection of u.i. random variables, therefore,

$$
\lim_{n \rightarrow \infty} VoI((x,a);\Fn,\Rm)=0
$$

For the case when $s$ is observed infinitely often, as shown in Theorem. \ref{consistencty}, $\P[a|\Rm] \rightarrow \delta_{a=a^{*}}$ as $m \rightarrow \infty$, therefore, 

\begin{align}
    \begin{split}
        G(\bx; \Rc^{\infty}, \Fn) &= \int_{A}\mu^{n}(\bx,\ba)\delta_{a=a*}da
    \end{split}\\
    \begin{split}
    &= \mu^{n}(\bx,\ba^{*})
\end{split}
\end{align}

Replacing $G(\bx; \Rc^{\infty}, \Fn)$ in $\VoI(s;\Rc^{\infty},\Fn)$ results in,

\begin{align}
\begin{split}
\lim_{m}\VoI(s; \Rm,\Fn)&= \lim_{m}\E_{r^{m+1}}\left[ \max_{x}G(x;\Rc^{m+1}, \Fn) \big| s, \Rm \right] - \max_{x}G(x;\Rm, \Fn)
\end{split}\\
\begin{split}
&= \E_{r^{\infty}}\left[ \max_{x}\mu^{n}(\bx,\ba^{*})\big| s, \Rc^{\infty} \right] - \max_{x}\mu^{n}(\bx,\ba^{*})
\end{split}\\
\begin{split}
&=  \max_{x}\mu^{n}(\bx,\ba^{*}) - \max_{x}\mu^{n}(\bx,\ba^{*})
\end{split}\\
\begin{split}
&=  0 
\end{split}\qed
\end{align}

\begin{ppr}
If $\lim_{n \rightarrow \infty} \VoI^{t}((x,a);\Rm,\Fn)= 0$ and $\lim_{m \rightarrow \infty} \VoI^{t}(s;\Rm,\Fn)= 0$ for all $(x,a)$ and $s$, then
$argmax_{x \in X}G(x;\Rm,\F^{\infty}) =argmax_{x \in X}\int_{A}\theta(x,a)\P[a|\Rm]da$ and $a^{*}$ is known
\label{reverese_xstar} 
\end{ppr}

\textit{Proof}

By Proposition \ref{existence_limit}, $\lim_{n \rightarrow \infty} \tilde{k}^{n}((x,a),(x,a)')=\tilde{k}^{\infty}((x,a),(x,a)')$ a.s for all $x,x' \in X$ and $a,a' \in A$. If the posterior variance $\tilde{k}^{\infty}((x,a),(x,a))=0$ for all $(x,a) \in X \times A$ then we know the global optimiser. Now, let's define $(\hat{x},\hat{a}) \in \hat{X} = \{x,a \in X \times A | \tilde{k}^{\infty}((x,a),(x,a))>0)\}$, then,

$$
\tilde{\Sigma}^{\infty}(x;(\hat{x},\hat{a}))=\frac{\int_{A}k^{\infty}((x,a),(\hat{x},\hat{a}) )\P[a|\Rm]da}{\sqrt{k^{\infty}((\hat{x},\hat{a}) ,(\hat{x},\hat{a}) )+\sigma_{\epsilon}^{2}}}>0
$$

Let's first assume $\tilde{\Sigma}^{\infty}(x_{1};(\hat{x},\hat{a}))\neq \tilde{\Sigma}^{\infty}(x_{2};(\hat{x},\hat{a}))$ for $x_{1},x_{2} \in X$. Then $VoI((x,a);\Rm\F^{\infty})$ must be strictly positive since for a value of $Z_{0} \in Z$,
$G(x_{1};\Rm \F^{\infty}) + \tilde{\Sigma}^{\infty}(x_{1};(\hat{x},\hat{a})) > G(x_{2};\Rm \F^{\infty}) + \tilde{\Sigma}^{\infty}(x_{2};(\hat{x},\hat{a}))$ for $Z>Z_{0}$ and vice versa. Therefore, $\tilde{\Sigma}^{\infty}(x''';(\hat{x},\hat{a}))= \tilde{\Sigma}^{\infty}(x'';(\hat{x},\hat{a}))$ must hold for any $x''', x'' \in X$ in order for $VoI((x,a))=0$, which results in,

$$
\frac{\int_{A}k^{\infty}((x''',a),(\hat{x},\hat{a}) )\P[a|\Rm]da}{\sqrt{k^{\infty}((\hat{x},\hat{a}) ,(\hat{x},\hat{a}) )+\sigma_{\epsilon}^{2}}} = \frac{\int_{A}k^{\infty}((x'',a),(\hat{x},\hat{a}) )\P[a|\Rm]da}{\sqrt{k^{\infty}((\hat{x},\hat{a}) ,(\hat{x},\hat{a}) )+\sigma_{\epsilon}^{2}}} 
$$

Since $\sigma_{\epsilon}^{2}>0$,

$$
\int_{A}\big[k^{\infty}((x''',a),(\hat{x},\hat{a}) )-k^{\infty}((x'',a),(\hat{x},\hat{a}) )\big]\P[a|\Rm]da = 0
$$

So $\tilde{\Sigma}^{\infty}(x;(\hat{x},\hat{a}))$ does not change for all $x \in X$. Moreover, by integrating with respect to $\hat{a}$, as $\tilde{K}(x;\hat{x}) = \int \tilde{\Sigma}^{\infty}(x'';(\hat{x},\hat{a})) d\hat{a}$ the resulting kernel does not vary with respect to $x$, it must be positive semidefinite, and symmetric. Therefore, by symmetry, the resulting  $\tilde{K}(x;\hat{x})$ does not change with respect to $\hat{x}$ and it must follow that the covariance matrix $\tilde{K}(x;\hat{x})$ is proportional to an all-ones matrix and the optimiser is known $argmax_{x\in X}G(x) = argmax_{x\in X}\int_{A}\theta(x,a)\P[a|\Rm]da$ but not necessarily its true value.\qed

The case when $\lim_{t \rightarrow \infty} \VoI(s;\Rm,\Fn)\rightarrow 0$ for all $s$ implies,

\begin{align}
\begin{split}
\E_{r^{m+1}}\left[ \max_{x}G(x;\Rc^{m+1}, \Fn) \big| s, \Rm \right] &= \max_{x}G(x;\Rm, \Fn)
\end{split}\\
\begin{split}
&= \E_{r^{m+1}}\left[ \max_{x}G(x;\Rm, \Fn)\big| s, \Rm \right]
\end{split}
\end{align}

Since both expectations are equal, it follows that $\max_{x}G(x;\Rc^{m+1}, \Fn) = \max_{x}G(x;\Rm, \Fn)$. Therefore,

$$
\max_{x}\int_{A}\mu^{n}(x,a)\P[a|\Rc^{m+1}]da = \max_{x}\int_{A}\mu^{n}(x,a)\P[a|\Rm]da
$$

Under some regularity conditions, as $m \rightarrow \infty$, the posterior distribution of $a$ approaches normality with mean $a^{*}$ and variance $mJ(a^{*})^{-1}$ where $a^*$ is the value that minimises the Kullback-Leibler divergence and $J$ is the Fisher Information.

\begin{align}
    \begin{split}
        KL(a) = \E\big[log\big(\frac{\P[r^{i}|a^{*}]}{\P[r^{i}|a]}\big)\big]
    \end{split}\\
    \begin{split}
        J(a)=-\E\big[\frac{d^{2}log\P(y|a)}{d a^{2}} \big |a \big]
    \end{split}
\end{align}

Therefore, variance reduces at rate $m^{-1}$. Equality for the posterior distribution at $m$ and $m+1$ must occur when both distributions are concentrated around $a^{*}$ as $\delta_{a=a*}$.\qed

However, it is possible that $\VoI(s;\Rm, \Fn)=0$ for a finite number of iterations. Particularly it would imply that the posterior \a distribution will not be affected by additional \rsNospace. Therefore, BICO would stop sampling from that specific \sNospace.

\begin{ppr}
$$
\VoI^t(s)=0 \Rightarrow \P[a|\Rc^{m+1}] = \P[a|\Rc^m]
$$
\label{prop:Voi_eq_zero_implies_pa_unchanged} 
\end{ppr}

\textit{Proof}

\begin{eqnarray}
\VoI^t(s) = \E_{r^{m+1}}\left[ \max_x \int_{a'} \mu^n(x,a')\P[a'|\Rc^{m+1}]da' \right] - \max_x \int_{a} \mu^n(x,a)\P[a|\Rc^{m}]da
\end{eqnarray}
Denote the current recommended solution as $x_r^t = \amax{x}\int_{a'} \mu^n(x,a)\P[a|\Rc^{m}]da$, and the $\VoI^t(s)$ can be rewritten as 
\begin{eqnarray}
0 
&=& \E_{r^{m+1}}\left[ \max_x \int_{a'} \mu^n(x,a')\P[a'|\Rc^{m+1}]da' \right] - \int_{a} \mu^n(x_r^t,a)\P[a|\Rc^{m}]da \\
&=& \E_{r^{m+1}}\left[ \max_x \int_{a'} \mu^n(x,a')\P[a'|\Rc^{m+1}]da'  - \int_{a} \mu^n(x_r^t,a)\P[a|\Rc^{m+1}]da \right] \\
&=& \E_{r^{m+1}}\left[ \max_x \int_{a'} \mu^n(x,a')-\mu^n(x_r^t,a')\P[a'|\Rc^{m+1}]da' \right]
\end{eqnarray}
Note that the random variable within the expectation is non-negative for all $r^{m+1}$. Since the expectation of the non-negative random variable is zero, every realisation of the random variable must be zero, for all $r^{m+1}$
\begin{eqnarray}
    \max_x \int_{a'} \mu^n(x,a')- \mu^n(x_r^t,a)\P[a'|\Rc^{m+1}]da'  = 0
\end{eqnarray}
If we denote the maximiser (which is a function of $r^{m+1}$) as $x_r^{t+1}(r^{m+1})$, the above equality may be written as 
\begin{eqnarray}
\int_{a'} \mu^n(x_r^{t+1}(r^{m+1}),a')- \mu^n(x_r^t,a)\P[a'|\Rc^{m+1}]da'  = 0.
\end{eqnarray}
Thus the above equality holds if
\begin{eqnarray}
    \mu^n(x_r^{t+1}(r^{m+1}),a') = \mu^n(x_r^t,a)
\end{eqnarray}
or equivalently $x_r^{t+1}(r^{m+1} ) = x_t^r$, the new maximiser \emph{does not depend} on $r^{m+1}$ and therefore
\begin{eqnarray}
    \max_{x} \int_{a'} \mu^n(x,a')\P[a'|\Rc^{m+1}]da'  
     &=&  \max_{x} \int_{a'} \mu^n(x,a')\P[a'|\Rc^{m}]da'
\end{eqnarray}
for all $r^{m+1}$ and for all $\mu^n$. The left hand side also does not depend on $r^{m+1}$ therefore $\P[a'|\Rc^{m+1}]$ does not depend on $r^{m+1}$ and we have that $\P[a'|\Rc^{m+1}] = \P[a'|\Rc^{m}]$. \qed

Therefore, BICO converges to finding the true \a $\ba^{*}$ and true optimal \x $\bx^{*}$ as t increases.

\subsection{BICO Relevance Determination}

In this section we show that if we use a squared exponential kernel then the hyperparameters determine the relevance of \ssNospace. Therefore,  non-relevant \ss will not be sampled by BICO.

\begin{remark}
Assuming a squared exponential kernel,

\begin{align}
\begin{split}
k^{0}((x,a)(x,a)')&=\sigma^{2}_{f}e^{-\frac{1}{2}\big (\frac{(x-x')^{2}}{l_{x}}+\frac{(a-a')^{2}}{l_{a}} \big ) }
\end{split}
\end{align}

and without loss of generality, a \a  $a \in A$, and a \x $x \in X$. Then $\VoI^t(s, \Rm,\Fn) = 0$ as $l_{a} \rightarrow \infty$. \label{noninfluentiala}
\end{remark}

\textit{Proof of Remark \ref{noninfluentiala}}

\vspace{1mm}

As $l_a \rightarrow \infty$ the posterior mean $\mu^n(\bx,\ba)$ only depends on the \x $x$. 

\begin{align}
\begin{split}
\lim_{l_a \rightarrow \infty}k^{0}((x,a)(x,a)') = \sigma^{2}_{f}e^{-\frac{1}{2}\big( \frac{(x-x')^{2}}{l_{x}} \big)} = k^{0}(x;x')
\end{split}\label{simplifiedk}
\end{align}

Let us denote $\tilde{X}_{x}^{n}=\{x^1,\dots,x^n\}$ and assume $\mu^{0}(x,a)=0$, then it follows from Equation (\ref{simplifiedk})
\begin{align}
    \begin{split}
\mu^n(x,a) &=-k^0((x,a),\tilde{X}^n)(k^0(\tilde{X}^n,\tilde{X}^n)+I\sigma)^{-1}Y^n
    \end{split}\\
    \begin{split}
    &= -k^0(x,\tilde{X}_{x}^n)(k^0(\tilde{X}_{x}^n,\tilde{X}_{x}^n)+I\sigma)^{-1}Y^n
    \end{split}\\
    \begin{split}
        &=\mu^n(x)
    \end{split}
\end{align}

Since $\mu^{n}$ does not depend on $a$, $G(x;\Rm, \Fn)= \mu^{n}(x)$, and the $VoI^{t}(\cdot)$ for $s \in \{1,\dots,N_{s}\}$ is

\begin{align}
\begin{split}
\VoI^t(\cdot) &= \VoI(s; \Rm, \Fn) 
\end{split}\\
\begin{split}
&= \E_{r^{m+1}}\left[ \max_{x}G(x;\Rc^{m+1}, \Fn) \big| s, \Rm \right] - \max_{x}G(x;\Rm, \Fn)
\end{split}\\
\begin{split}
&= \E_{r^{m+1}}\left[ \max_{x}\mu^{n}(x)\big| s, \Rm \right] - \max_{x}\mu^{n}(x)
\end{split}\\
\begin{split}
&=  \max_{x}\mu^{n}(x) - \max_{x}\mu^{n}(x)
\end{split}\\
\begin{split}
&=  0
\end{split}\qed
\end{align}

Therefore, external data is never collected if $a$ is not "influential" on the predicted simulation output $\mu^{n}(x,a)$.

\subsection{Implementation Details}

In this section we consider implementation details that have omitted from the main document for brevity. 

For the Gaussian process, the hyperparameters are found by maximising the marginal likelihood using the L-BFGS-B algorithm with several restarts. This is repeated for every iteration of the BICO algorithm. 

The Value of Information (\VoI) for a \xaNospace, as mentioned in Section~\ref{VoI_KG}, can be computed by the traditional Knowledge Gradient for Continuous Parameters (\citeauthor{Frazier209}, \citeyear{Frazier209}). More specifically, we discretise set $X$ by Latin hypercube sampling (LHS) but including each fantasised sample $(\bx,\ba)^{n+1}$ in the discretisation. For  $A$, we sample from the posterior distribution to obtain a discrete set. Then, to compute the VoI of sampling the simulator according to Equation~\ref{VoI},
we replace $G(\bx; \Rm, \Fn)$ and $\tilde{\Sigma}^{n}(\bx;(\bx,\ba)^{n+1})\}$ by their Monte-Carlo estimates using $N_{A}$ samples from $a_{i} \sim \P[a|\Rc^{m}]$,

\begin{align}
\begin{split}\label{eq:eq10}
G(\bx; \Rm, \Fn) \approx \frac{1}{N_{A}}\sum_{a_{i}\in A_{MC}}\mu^{n}(\mathbf{x},\mathbf{a}_{i})
\end{split}
\end{align}

\begin{align}
\begin{split}
\tilde{\Sigma}^{n}(\bx;(\bx,\ba)^{n+1})\} \approx  \frac{1}{N_{A}}\sum_{a_{i}}\tilde{\sigma}^{n}((\mathbf{x},\mathbf{a}_{i});(\mathbf{x},\mathbf{a})),
\end{split}
\end{align}

which is then optimised using the Nelder-Mead optimiser with several restarts. Similarly, the \VoI\text{ }of sampling the \s can be estimated by its Monte-Carlo approximation by marginalising over the \a distribution $\P[a|\mathscr{R}^{m+1}]$ and predicted \r distribution $r_{i}^{m+1}\sim \P[r|s,\ba^*]$,
  
  \begin{align}
  \begin{split}
  \VoI(s; \Rm, \Fn) \approx \frac{1}{N_{r}N_{A}}\sum_{r^{m+1}_{i}}\sum_{a_{i}}[ \mu(\mathbf{x}_{r}(\mathscr{D}^{m+1}),\mathbf{a}_{i})- \mu(\mathbf{x}_{r}(\mathscr{D}^{m}),\mathbf{a}_{i})]
  \end{split}\label{DlossMC}
  \end{align}
  
  However, $\P[a|\mathscr{R}^{m+1}]$ would have to be updated for each value $r_{i}^{m+1}$ and generate new Monte-Carlo samples from  $\P[a|\mathscr{R}^{m+1}]$. Therefore, we implemented the following modification using importance sampling weights where samples from $\P[a|\mathscr{R}^{m}]$ and $r_{i}^{m+1}\sim \P[r|s,\ba^*]$ are produced instead.
   
     \begin{align}
   \begin{split}
   \VoI(s; \Rm, \Fn) \approx \frac{1}{N_D N_{A}}\sum_{r^{m+1}_{i}}\sum_{a_{i}}[ \mu(\mathbf{x}_{r}(\mathscr{D}^{m+1}),\mathbf{a}_{i})- \mu(\mathbf{x}_{r}(\mathscr{D}^{m}),\mathbf{a}_{i})]\frac{\mathbb{P}[\mathbf{a}|\mathscr{D}^{m+1})]}{\mathbb{P}[\mathbf{a}|\mathscr{D}^{m}]}
   \end{split}
   \end{align}
   
   which allows to generate the Monte-Carlo samples just once for each $\VoI(s; \Rm, \Fn)$ estimation.
   
\end{document}